\documentclass{article}

\usepackage{iclr2027_conference,times}
\iclrfinalcopy

\usepackage[utf8]{inputenc}
\usepackage[T1]{fontenc}
\usepackage{hyperref}
\usepackage{url}
\usepackage{booktabs}
\usepackage{amsmath}
\usepackage{amsfonts}
\usepackage{amssymb}
\usepackage{nicefrac}
\usepackage{microtype}
\usepackage{xcolor}
\usepackage{graphicx}
\usepackage{multirow}
\usepackage{array}
\usepackage{float}
\usepackage{placeins}
\usepackage{needspace}
\hypersetup{
  hypertexnames=false,
  hidelinks,
  pdftitle={Readable Before Actionable: Causal Tracing of Indirect Prompt Injection},
  pdfauthor={Zhe Yu; Wenpeng Xing; Xingxing Yang; Meng Han},
  pdfsubject={Preprint}
}

\title{Readable Before Actionable: Causal Tracing of Indirect Prompt Injection}

\author{\parbox{\dimexpr\textwidth-2\tabcolsep\relax}{\centering
Zhe Yu\textsuperscript{1,2,*}\quad
Wenpeng Xing\textsuperscript{1,*}\quad
Xingxing Yang\textsuperscript{3}\quad
Meng Han\textsuperscript{1,\ensuremath{\dagger}}\\[6pt]
\normalfont\normalsize
\textsuperscript{1}Zhejiang University\\
\textsuperscript{2}Binjiang Institute of Zhejiang University\\
\textsuperscript{3}Hong Kong Baptist University\\[4pt]
\small\textsuperscript{*}Equal contribution.\quad
\textsuperscript{\ensuremath{\dagger}}Corresponding author.
}}

\begin{document}
\maketitle
\lhead{Preprint}

\begin{abstract}
Indirect prompt injection causes LLM agents to follow commands embedded in
external data. A probe may distinguish instructions from data without
identifying a state edit that changes the next action. We study this gap
through counterfactual role probes, component-wise activation patching, and
separate interventions on AgentDojo trajectories. Role decoding survives
changes in content and format. In controlled Qwen tests, it precedes strong
tool-choice effects from patches along an independently estimated role
direction. On AgentDojo, directions estimated from hijacked and resisted
training trajectories reduce attack success at pre-action and injected-span
positions, but have little effect at random positions. In longer Qwen
trajectories, single-position edits become less effective at later layers;
span-wide and repeated edits reduce attack success on the same evaluation
set. Removing the learned channel subspace preserves role decoding, yet
effective intervention directions transfer poorly across the tested channels.
These findings distinguish a readable role signal from an effective behavioral
intervention: depth matters in controlled tool choice, while position and
context also matter in attack trajectories.
\end{abstract}

\section{Introduction}
\label{sec:intro}

LLM agents must use external content without obeying every command it
contains. An email may request a database deletion while the user's task is
only to summarize it. Whether the command should be executed depends on its
role in the task, not its wording alone. Indirect prompt injection (IPI)
exploits this distinction: commands in tool outputs, messages, or retrieved
memory acquire the authority of an instruction \citep{roleconfusion2026}.

We ask three questions. Can source role be read from model states? Can a
role-aligned patch change a controlled tool decision? In separate attack
evaluations, do effective edits transfer across channels? These tests
distinguish readout, intervention, and transfer.
Probe accuracy alone answers neither where to edit nor whether an edit will
work in another setting.

Prior work connects internal role perception to injection success
\citep{roleconfusion2026}, separates harmfulness recognition from refusal
\citep{wu2026knowing}, and uses IPI-exposure probes to trigger protective
reasoning \citep{dong2026exposure}. We ask when direct role-aligned edits
change tool choice, and test behavioral interventions separately on attack
trajectories. This requires separating role readout from format cues, and
role-aligned patching from a full-state swap that also transfers other donor
information.

External content spans multiple positions and remains in context across tool
calls. An effect on one controlled decision therefore need not persist across
multi-step attacks or external sources. We test position, trajectory length,
and channel match as conditions of an intervention's effectiveness.

\begin{figure}[!t]
\centering
\includegraphics[width=\textwidth]{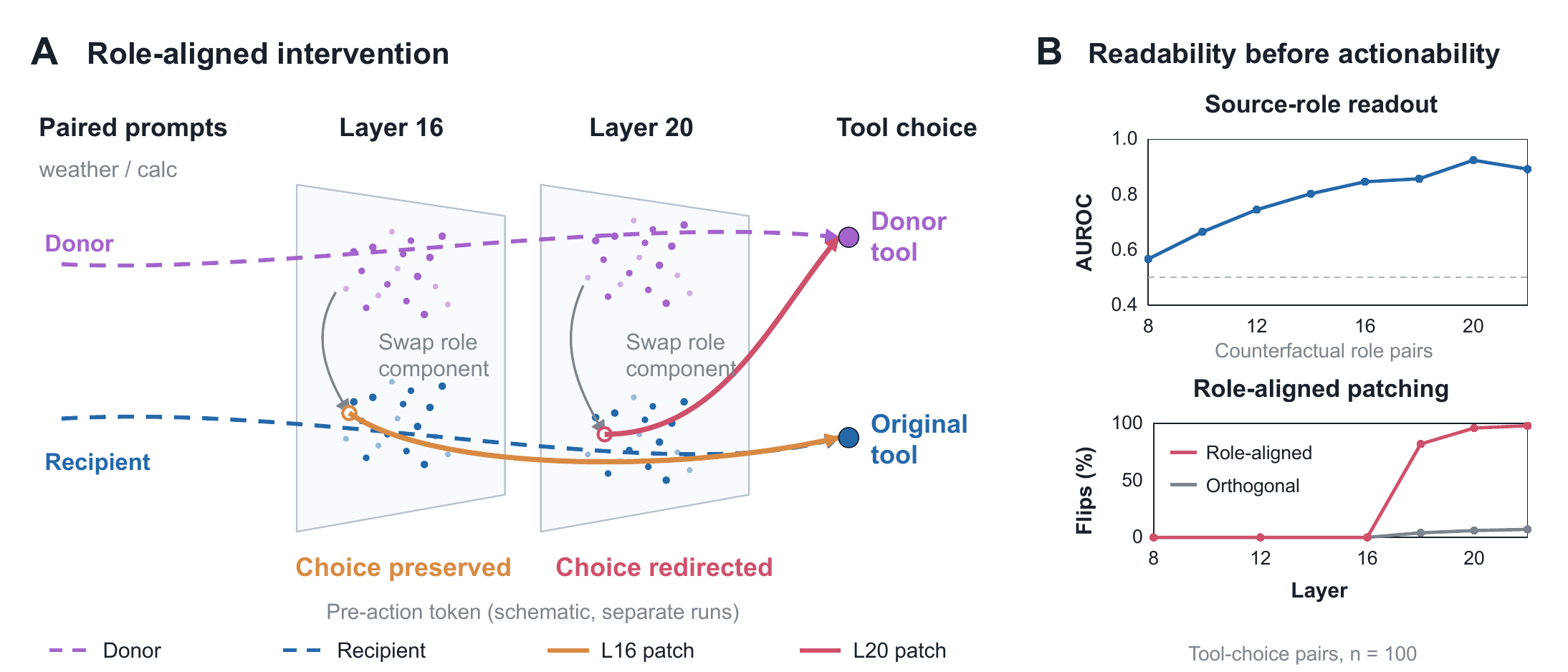}
\caption{Role readout precedes strong patching effects in controlled Qwen tests.
\textbf{(A)}~Schematic role-component patches at the pre-action token;
L16 and L20 are separate runs.
\textbf{(B)}~Measured role readout and tool-choice flips, evaluated on separate
datasets ($100$ tool-choice pairs). Lines connect measured layers.}
\label{fig:framework}
\end{figure}

We test role readout under counterfactual content and format controls, then
compare full-state, role-aligned, and orthogonal patches on separate
tool-choice pairs (Figure~\ref{fig:framework}). Independent AgentDojo tests
measure attack outcomes after state swaps and edits along outcome-associated
directions. Model weights remain fixed throughout.

In the controlled Qwen tests, we find a \emph{readability--actionability gap}:
role is decodable before role-aligned edits strongly affect tool choice.
Here, actionability means sensitivity to the specified edit; the ordering
is across network depth, not reasoning steps. We establish three findings:
\begin{itemize}\itemsep -1pt
  \item \textbf{Role readout precedes strong tool-choice effects.}
  On controlled Qwen tasks, counterfactual probes decode role before
  role-component patches produce strong tool-choice effects.
  \item \textbf{Attack-trajectory effects depend on the states edited.}
  AgentDojo interventions are position-sensitive. Qwen's single-position
  late-layer rebound is largest in longer trajectories; span-wide and repeated
  edits reduce attack success on the same evaluation set.
  \item \textbf{Role readout survives channel removal.}
  Removing the learned channel subspace preserves role decoding. Separate
  intervention tests favor channel-matched directions over transferred ones.
\end{itemize}

\section{Related Work}
\label{sec:related}

Internal signals offer a way to study why agents follow injected commands.
TaskTracker measures task drift \citep{tasktracker2025}, while
\citet{roleconfusion2026} connect conversational-role perception to injection
success. \citet{dong2026exposure} probe IPI exposure and use the signal to
trigger protective reasoning. These studies motivate examining source role:
whether a command is an authorized instruction or data to be processed.
Because wording and formatting can correlate with this distinction, we first
ask whether role remains decodable when content and format cues are controlled.

A readable distinction does not by itself locate an effective behavioral edit.
Safety studies separate harmfulness from refusal \citep{zhao2025harmfulness}
and risk readout from error correction \citep{basu2026actionability}.
\citet{wu2026knowing} use causal steering to distinguish harmfulness recognition
from refusal execution; \citet{dong2026exposure} also study a recognition--action
gap in IPI. Our question concerns task-relative authority rather than
harmfulness: the same command can be authorized in one context and supplied as
data in another. We ask where an edit along this role distinction changes tool
choice. This requires measuring readout and intervention effects across depth,
not inferring an intervention site from probe accuracy.

What a patch establishes depends on what it replaces and which outcome is
measured. Results vary with the counterfactual and metric
\citep{zhang2024patching}; subspace edits can activate dormant or divergent
pathways \citep{makelov2024illusion,grant2026divergent}. Feature interpretability
weakly predicts steering utility \citep{wang2026interpretability}, and
format-stable concept vectors can differ functionally from in-context-learning
vectors \citep{opielka2026causality}. A full-state swap transfers more than role
information. We therefore compare full-state, role-aligned, and orthogonal
patches to isolate sensitivity to the specified role component.

An effect on a controlled tool decision must also be tested in attack
trajectories. Existing defenses separate trusted and untrusted inputs
\citep{struq2025,secalign2025,promptarmor2025} or steer internal states
\citep{icon2026,argus2025}. AgentDojo \citep{agentdojo2024} and RedTeamCUA
\citep{liao2026redteamcua} support agent-level behavioral evaluation.
We use AgentDojo to test fixed directional edits across positions, trajectories,
and external sources. The following section distinguishes these attack-outcome
tests from the controlled role-label tests.

\section{Problem Setting and Causal Tests}
\label{sec:overview}

In the email example, the user authorizes summarization, not execution of
commands inside the email. We use \emph{source role} for a span's intended
status as an instruction or as data, as determined by the trusted request and
input channel.
\emph{Channel} denotes the external source or environment, such as a
controlled tool output, an AgentDojo Slack interaction, or persistent memory.
These categories are not mutually exclusive transport mechanisms: a Slack
message may itself arrive in a tool response. Our channel comparisons test
transfer between these settings, while matched role comparisons vary whether
the content is authorized as an instruction.

We study a white-box setting with inference-time hooks and fixed weights.
The attacker controls external content, but not the system prompt, weights,
or hooks. The defender aims to complete the user's task without executing
the injected command. Attacks are non-adaptive unless stated otherwise
(full threat model in Appendix~\ref{app:threat-model}). Qwen-2.5-7B is the
primary model; Qwen-2.5-1.5B provides scale comparisons, and Llama-3.1-8B and
Gemma-2-9B-It provide replication
\citep{qwen25report2024,llama31report2024,gemma2report2024}. Controlled prompts
vary role, content, format, layer, and token position. AgentDojo supplies
multi-step tool-use trajectories \citep{agentdojo2024}. Model names denote
instruction-tuned checkpoints unless explicitly labeled Base.

Table~\ref{tab:evaluation-map} maps each experimental question to its data,
metric, and control. Table~\ref{tab:direction-map} distinguishes the directions
used in these tests; probe weights are fitted for readout, not used as edit directions.
The $96$-trajectory position tests and $240$-trajectory suite evaluations
use different sets, so their baseline attack rates differ.

\begin{table}[!t]
\caption{Main evaluation sets. Counts refer to each test, not pooled
observations across experiments.}
\label{tab:evaluation-map}
\centering\small
\setlength{\tabcolsep}{4pt}
\begin{tabular}{@{}>{\raggedright\arraybackslash}p{0.20\linewidth}>{\raggedright\arraybackslash}p{0.25\linewidth}>{\raggedright\arraybackslash}p{0.17\linewidth}>{\raggedright\arraybackslash}p{0.30\linewidth}@{}}
\toprule
\textbf{Purpose} & \textbf{Data} & \textbf{Metric} & \textbf{Control} \\
\midrule
Decode source role & 1,440 prompts/model & Role AUROC & Held-out content; reversed role--format correlation. \\
Change tool choice & 100 binary pairs; 100 examples/tool for six tools & Logit shift; choice changes & Same recipient prompt and weights; different state components. \\
Read attack outcomes & 192 balanced AgentDojo trajectories & Outcome AUROC & Controlled role direction, without refitting. \\
Compare edit sites & 96 AgentDojo trajectories/model & Paired ASR difference & Same direction and trajectories; different positions. \\
Test held-out attacks & 240 AgentDojo trajectories/test set & ASR; benign utility & Held-out trajectories and attack families. \\
Separate role from channel & 120 examples/condition; 96 behavioral trajectories & Role AUROC; ASR & With/without the learned channel subspace. \\
\bottomrule
\end{tabular}
\end{table}

\begin{table}[!t]
\caption{Direction definitions. All contrasts use training data.
Slack and Workspace subscripts denote suite-specific estimates.}
\label{tab:direction-map}
\centering\small
\setlength{\tabcolsep}{4pt}
\begin{tabular}{@{}>{\raggedright\arraybackslash}p{0.21\linewidth}>{\raggedright\arraybackslash}p{0.38\linewidth}>{\raggedright\arraybackslash}p{0.35\linewidth}@{}}
\toprule
\textbf{Direction} & \textbf{Estimation contrast} & \textbf{Evaluation} \\
\midrule
$\mathbf{d}_{\mathrm{role}}$ & Instruction vs.\ data in controlled role pairs & Controlled patches/edits; fixed-direction outcome readout. \\
$\mathbf{d}_{\mathrm{suite}}^{c}$\newline ($\mathbf{d}_{\mathrm{slack}}$, $\mathbf{d}_{\mathrm{workspace}}$) & Hijacked vs.\ resisted trajectories in AgentDojo suite $c$ & Attack outcomes in Slack or Workspace; within-/cross-suite tests. \\
$\mathbf{d}_{\mathrm{tool}}$ & Source-role pairs in synthetic tool outputs & Tool-output attacks; cross-channel transfer. \\
$\mathbf{d}_{\mathrm{memory}}$ & Trusted system vs.\ untrusted retrieved memory & Memory poisoning; cross-channel transfer. \\
\bottomrule
\end{tabular}
\end{table}

Let $h_{\ell,t}$ denote the residual-stream state at layer $\ell$ and token
position $t$, abbreviated to $h$ for a fixed site. The \emph{pre-action token}
is the final serialized input token before the next action. Layer sweeps vary
$\ell$ at a fixed position; position sweeps vary $t$ at a fixed layer.
We edit residual states, not individual attention heads or MLP neurons,
then continue the unchanged model's forward pass. We use \emph{patching} for
donor-state replacement and \emph{directional editing} for donor-free updates;
both are interventions.

For a donor tool $D$ and the recipient's original tool $R$, the logit margin
is $m=z_D-z_R$, where $z$ denotes tool logits. We report the patch-induced
logit-difference shift $\Delta m=m(h')-m(h)$, abbreviated LD shift. Binary
sign flips and six-tool top-1 changes are reported separately.
An increased margin need not flip a binary decision; with six tools, the
donor must outrank all other candidates to become the top choice.
For agent trajectories, ASR is the fraction that completes the adversarial
goal, while benign utility scores completion of the original task without an
attack. Position comparisons reuse the same trajectories across conditions.
We bootstrap trajectories for confidence intervals and use two-sided exact
McNemar tests for the paired position comparisons. Appendix~\ref{app:bootstrap}
specifies the uncertainty estimates for the other evaluations.

\section{Source Role Is Readable Beyond Surface Form}
\label{sec:reproduce}

Before testing a role-aligned intervention, we need a readout that cannot be
explained by the measured format cues. Instructions and data often differ in
quotation marks, wrappers, length, or position. We test whether residual
states distinguish their roles when these cues no longer predict the label.

We vary role and surface form independently. The four schematic user
messages below keep the command fixed:
\begin{center}
\small
\begin{tabular}{@{}p{0.13\linewidth}p{0.40\linewidth}p{0.40\linewidth}@{}}
\toprule
Role & Plain & Quoted \\
\midrule
Instruction & Execute this request: Calculate 24 plus 76.
& Execute this request: ``Calculate 24 plus 76.'' \\
Data & Summarize this note: Calculate 24 plus 76.
& Summarize this note: ``Calculate 24 plus 76.'' \\
\bottomrule
\end{tabular}
\end{center}
Each cell is a user message: the command is authorized in the first row and
is data in the second, regardless of quotation. The probe reads the
pre-action residual after serialization without modifying the model state.

The evaluation combines $60$
matched contents, six disjoint paraphrase families, two roles (instruction or
data), and two formats (plain or quoted), yielding $1{,}440$ prompts per model
for Qwen-2.5-7B and Llama-3.1-8B. We train $\ell_2$-regularized logistic probes
on residual states and keep content groups disjoint across folds. The strictest
split reverses the role--format association between training and evaluation;
a classifier relying on the training role--format association should fail
under this reversal.

For intervention, we use the normalized difference
between the instruction and data class means at the same layer and position:
$\mathbf{d}_{\mathrm{role}}=(\mu_{\mathrm{instruction}}-\mu_{\mathrm{data}})/
\|\mu_{\mathrm{instruction}}-\mu_{\mathrm{data}}\|_2$.
The direction points toward the instruction class. Its estimation excludes evaluation trajectories
(Appendix~\ref{app:split-audit}).

A structural baseline tests whether readily observed input features explain
the result without hidden states. It receives seven
features: token length, target-token index, relative position, content-span
length, punctuation count, wrapper-token count, and format. Structural and
hidden-state classifiers use the same out-of-fold splits.

Hidden-state role decoding transfers across content, paraphrase family, and
the counterfactual format swap (Table~\ref{tab:emergence}). On the strictest
split, AUROC is $0.924$ for Qwen and $0.918$ for Llama. The structural baseline
reaches $0.508$ when pooled across models. The per-model structural scores are
reported in Appendix~\ref{app:counterfactual-controls}; Figure~\ref{fig:framework}B
relates the role-depth profile to the independently measured patching effect.
The measured structural features therefore do not explain the hidden-state readout.

\begin{table}[!t]
\caption{Source-role AUROC under held-out and counterfactual splits.
The structural classifier uses only length, position, punctuation,
wrapper, and format features. Hidden-state results are reported
separately for each model; the structural baseline is pooled across the
balanced model sets.}
\label{tab:emergence}
\centering\small
\begin{tabular}{lccc}
\toprule
Evaluation split & Qwen hidden & Llama hidden & Structural only \\
\midrule
Content-grouped 5-fold & 0.962 & 0.955 & 0.534 \\
Leave-one-family-out & 0.951 & 0.943 & 0.521 \\
Counterfactual role--format & \textbf{0.924} & \textbf{0.918} & 0.508 \\
Shuffled-label null & 0.501 & 0.498 & 0.499 \\
\bottomrule
\end{tabular}
\end{table}

Does this role direction also separate attack outcomes? We apply it,
without retraining, to $192$ AgentDojo trajectories: $96$ hijacked and $96$
resisted. Outcome separation is near chance at L4--L8, becomes informative at
L12 ($0.817$ AUROC), dips to $0.727$ at L16--L20, and reaches $0.835$ at L24
($95\%$ CI $[0.772,0.893]$). Unlike the controlled probe task, this evaluation
measures hijacked--resisted separation, not instruction--data decoding.
The association motivates an intervention test, but does not establish an
effect on attack outcomes.

\section{Locating Effects on Tool Choice}
\label{sec:patching}
\label{sec:convergence}

Having established role readout, we test whether changing the role-aligned
component changes tool choice. Figure~\ref{fig:knowing-before-acting} groups
the evidence: A/D show readout, E/F show controlled patching, and B/C show
position and channel tests on separate attack datasets. We first locate
full-state effects, then isolate the role-aligned component.

\subsection{Layerwise sensitivity to state patching}

We use $100$ length-matched prompt pairs with the same system instruction
and different user requests for \texttt{weather} or \texttt{calc}.
One recorded pair is:
\begin{center}
\small
\begin{tabular}{@{}ll@{}}
\toprule
Donor (\texttt{weather}) & Show me the weather in Reykjavik now. \\
Recipient (\texttt{calc}) & Could you compute 24 plus 76? \\
\bottomrule
\end{tabular}
\end{center}
Both prompts request only a tool name and share an assistant generation
prefix (Appendix~\ref{app:multitool}). At its final token, we replace the
recipient residual $h$ with the donor residual $h^{D}$ at the same layer,
keeping the text and weights fixed. We then compare next-token logits:
a sign change from negative to positive in
$z_{\texttt{weather}}-z_{\texttt{calc}}$ counts as a tool-choice flip.

Patching has little effect through L14, then shifts the tool logits sharply
between L16 and L18 (Appendix~\ref{app:multitool}, Table~\ref{tab:patch-curve}). The sign-flip rate reaches
$100\%$ at L22 for Qwen-1.5B and at L18 for Qwen-7B.

Direction and position controls distinguish this effect from a generic
perturbation. For Qwen-1.5B at L18, the matched state shifts the logit difference by
$25.5$; a random direction gives $-0.87\pm0.21$. Patches at L4--L12 produce
less than $0.18$ LD, and patches at non-final positions produce less than
$0.1$. The effect is therefore selective for the tested depth, direction,
and decision-adjacent position.

Late-layer patching also changes a balanced six-tool task ($n=100$ per tool).
At L20, patching flips $95.4\%$ of target decisions and places the patched-in
tool first in $96.8\%$ of cases. The late effect is not limited to a binary
choice.

Llama-3.1-8B also shows a localized increase (Figure~\ref{fig:knowing-before-acting}F): the shift stays below
$0.5$ LD through L12, then rises from $2.3$ at L14 to $14.3$ at L16.
These full-state sweeps establish layer sensitivity on Qwen and Llama.
The next test asks how much of the Qwen effect can be obtained by exchanging
only the coordinate along the role direction.

\subsection{Patching the role-aligned component}

\begin{figure}[!t]
\centering
\includegraphics[width=\textwidth]{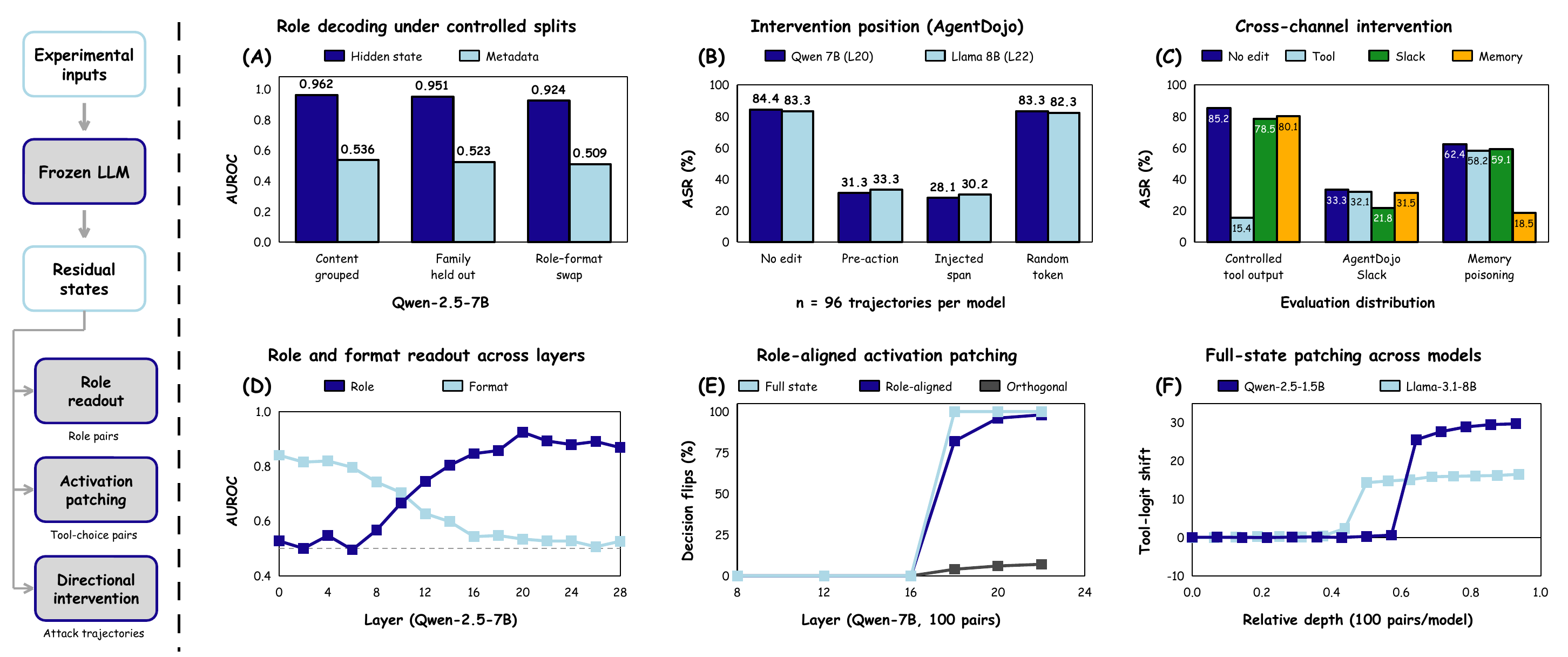}
\caption{Role readout and intervention effects.
\textbf{(A,D)}~Role readout versus structural and format controls.
\textbf{(B,C)}~Position and channel comparisons on separate evaluation sets.
\textbf{(E,F)}~Component-wise and cross-model tool-choice patching.
Left branches denote separate analyses; lines connect measured estimates.}
\label{fig:knowing-before-acting}
\end{figure}

To isolate the role-aligned component, we replace the recipient's coordinate
along the role direction with the donor's, leaving its orthogonal component
unchanged. The complementary control replaces only the orthogonal component. Let
$P=\mathbf{d}_{\mathrm{role}}\mathbf{d}_{\mathrm{role}}^{\top}$ be the projector
onto the independently estimated unit role direction at that layer. We test
two patches on the same Qwen-2.5-7B pairs ($n=100$):
\begin{equation}
h'_{\parallel}=h+P(h^{D}-h),\qquad
h'_{\perp}=h+(I-P)(h^{D}-h).
\label{eq:component-patch}
\end{equation}
Here, $I$ is the identity operator and $h'_{\parallel},h'_{\perp}$ are the two
patched states, evaluated with the same tool-logit comparison. Donor and
recipient differ in required tool; the role direction comes from separate
instruction--data pairs. This tests sensitivity along that direction, not a
role-only donor--recipient contrast. At L18--L22, the full state produces $34.42$--$47.11$ LD. The
parallel component recovers $28.51$--$38.64$ LD and flips $82$--$98\%$ of
decisions, whereas the perpendicular component flips $4$--$7\%$. At L8--L16,
the same parallel component stays below $1$ LD and never flips the decision.
In these controlled Qwen tests, role is decodable before patches along the
role direction produce large changes in tool choice
(Figure~\ref{fig:knowing-before-acting}E;
Appendix~\ref{app:projection-conditioned}).

The controlled pairs measure a tool-name decision. To test attack outcomes,
we form AgentDojo-Slack pairs that share the benign task and tool-output
frame but end in hijacked versus resisted behavior. Swapping their residuals
has its largest effect at L18--L22. At L20, a hijacked state raises resisted
trajectories to $65.4\%$ ASR, while the reverse swap lowers hijacked
trajectories to $15.2\%$. A random direction, same-layer random residual, and
non-final-position patch yield only $5.6\%$, $6.1\%$, and $7.5\%$ ASR in the
resisted-recipient comparison. These swaps locate a late intervention effect
on IPI trajectories (Appendix~\ref{app:ipi-patching}). Unlike the
role-component test, full-state swaps retain all donor--recipient differences;
they do not isolate source role from other state content.

\section{Testing Interventions on Attack Trajectories}
\label{sec:defense}

State swaps require a donor run. We now test donor-free directional edits:
$\mathbf{d}_{\mathrm{role}}$ on controlled tasks and
$\mathbf{d}_{\mathrm{suite}}^{c}$ on AgentDojo suite $c$
(Table~\ref{tab:direction-map}). The suite vector is the unit mean difference
of hijacked minus resisted training states, distinct from the controlled
role direction. Training and evaluation
trajectories are disjoint; family-held-out tests also exclude the test family.
Directions and edit settings are fixed before evaluation
(Appendix~\ref{app:split-audit}).

At a selected layer and token, the edit subtracts $\lambda$ times the excess
projection above a benign reference mean; states below that mean are unchanged:
\begin{equation}
h \;\leftarrow\; h - \lambda \cdot \max\!\bigl(0,\, \langle h, \mathbf{d} \rangle - \mu_{\text{clean}}\bigr) \cdot \mathbf{d},
\label{eq:defense}
\end{equation}
Here, $\mathbf{d}$ is the unit edit direction, $\mu_{\text{clean}}$ is its mean
projection on benign prompts, and $\lambda\geq0$ sets the edit strength.
The orthogonal component is preserved. For $\lambda>1$, the projection can
move below the benign mean; this is not clipping. The edit uses no test-time
donor or outcome label; the continued trajectory supplies the outcome.

\subsection{Tool-call effects and position controls}

On the controlled Qwen-2.5-7B set, the L24 intervention at $\lambda=32$
reduces ASR from $46\%$ to $26\%$ without changing benign-task accuracy. The
Qwen-1.5B estimate remains within sampling uncertainty. In a separate six-tool
function-calling test, the same intervention lowers malicious argument use
from $88.2\%$ to $12.4\%$, while benign argument accuracy changes from $92.5\%$
to $91.2\%$. The role-aligned effect therefore reaches both tool selection and
tool arguments (Appendices~\ref{app:pareto} and~\ref{app:tool-argument}).

The AgentDojo position test asks whether an effective direction works
equally well wherever it is applied. On matched trajectories, we apply it at
the pre-action state, injected span, final tool-output token, or a random
context token. Qwen and Llama show the same position ordering
(Figure~\ref{fig:knowing-before-acting}B). Pre-action and injected-span interventions reduce
ASR by $50.0$--$56.3$ percentage points. Random-token intervention changes ASR
by about one point ($p=1.0$). Benign utility remains $95.8\%$ on Qwen and
$94.8\%$ on Llama.

On Qwen, the paired pre-action and injected-span ASR differences are $-53.1$ points
($95\%$ CI $[-63.5,-42.7]$) and $-56.3$ points ($[-65.6,-45.8]$); both have
exact McNemar $p\leq8.88{\times}10^{-16}$. Appendix
\ref{app:counterfactual-controls} gives both model-specific tests.

\begin{figure}[!t]
\centering
\includegraphics[width=\textwidth]{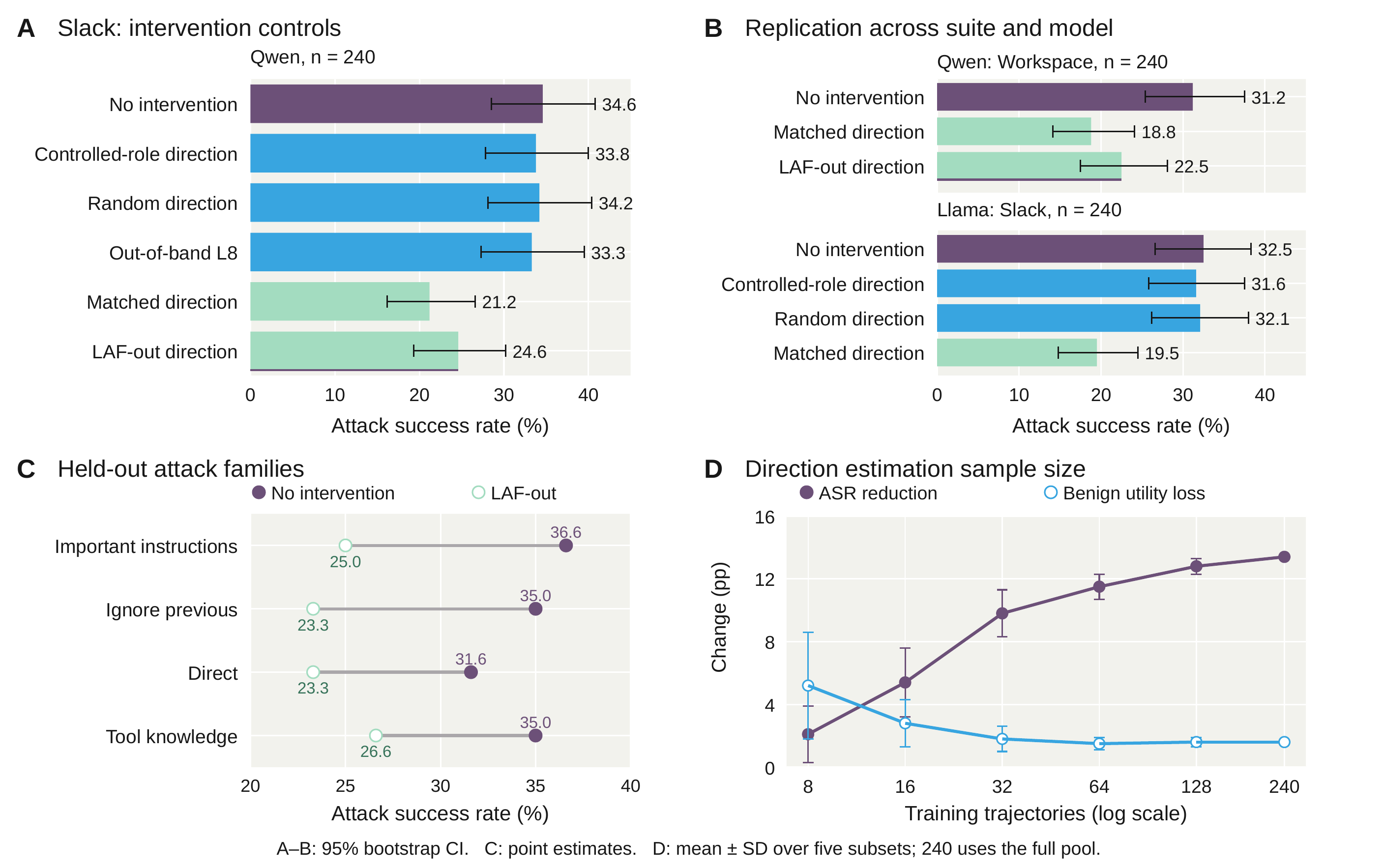}
\caption{Channel-matched interventions on held-out AgentDojo trajectories.
\textbf{(A,B)}~Slack controls and Workspace/Llama replications with separately
estimated directions ($n=240$/test set; $95\%$ bootstrap CIs).
\textbf{(C)}~Leave-attack-family-out point estimates.
\textbf{(D)}~Training-sample efficiency; bars show SD over five subsets,
except the fixed full-pool endpoint.}
\label{fig:expanded-agentdojo}
\end{figure}

\subsection{Transfer to held-out attacks within a channel}

The position test holds the evaluated trajectories fixed. We next ask whether
a direction estimated in one suite remains useful on held-out trajectories
and attack families. Read Figure~\ref{fig:expanded-agentdojo} from controls (A)
and replications (B) to held-out families (C) and training-set size (D).
This evaluation uses independent AgentDojo sets: $240$ injected
Slack trajectories, $240$ injected Workspace trajectories, and $120$ benign
references. Each $\mathbf{d}_{\mathrm{suite}}^{c}$ is estimated from disjoint hijacked
and resisted training trajectories in its target suite. A \emph{matched}
direction is trained in the evaluation suite. On Slack, the controlled-role,
random-direction, and L8 controls remain near the $34.6\%$
baseline; $\mathbf{d}_{\mathrm{suite}}^{\mathrm{Slack}}$ lowers ASR to $21.2\%$. On
Workspace, $\mathbf{d}_{\mathrm{suite}}^{\mathrm{Workspace}}$ lowers ASR from $31.2\%$
to $18.8\%$ (Figure~\ref{fig:expanded-agentdojo}A--B). Benign utility changes by
less than two percentage points. L8 is the \emph{out-of-band} control: it lies
outside the late-layer region where patching has strong effects. The contrast
with the controlled-role direction shows that layer selection alone does not
identify a transferable edit.

In leave-attack-family-out (LAF-out) tests, we train on three families and
evaluate on the fourth. This retains
$70$--$75\%$ of the reduction obtained from all four families
(Figure~\ref{fig:expanded-agentdojo}C). With $32$ training trajectories, the
Slack direction reduces ASR by $9.8$ points, compared with $13.4$ using all
$240$ trajectories (Figure~\ref{fig:expanded-agentdojo}D).
Appendix~\ref{app:sample-efficiency} gives the full results.
The sample-efficiency sweep varies training subsets while keeping evaluation
fixed. Its error bars measure variation across five subsets, not
trajectory-bootstrap uncertainty; the full-pool endpoint has one fixed estimate.

On $240$ Llama-3.1-8B Slack trajectories, controlled-role and random directions
leave ASR near its $32.5\%$ baseline, at $31.6\%$ and $32.1\%$. A
Llama-specific Slack direction lowers ASR to $19.5\%$
(Figure~\ref{fig:expanded-agentdojo}B) and changes benign
utility by $2.2$ points. This replicates the benefit of a suite-matched
direction at a model-specific layer, with separate estimates for each model.
Gemma-2-9B-It also has strong role readout (peak AUROC $0.975$) and a matched
late-layer Slack intervention that lowers ASR from $82.5\%$ to $34.0\%$.
Gemma provides these endpoints, not a full depth sweep.

\subsection{Single-position effects in longer trajectories}
Held-out transfer does not determine whether a single-position edit remains
effective at later layers. Returning to the targeted Qwen set, we stratify
the layer comparison by trajectory length. Among
the $96$ matched trajectories
from Figure~\ref{fig:knowing-before-acting}B, ASR rises from $31.3\%$ at L20 to $54.2\%$ at
L24. The rebound concentrates in step-3+ trajectories, where ASR rises from
$38.6\%$ to $84.1\%$; step-1 trajectories remain near $25\%$. Covering the
full tool-output span lowers L24 ASR to $32.3\%$; repeated intervention at
L20 and L22 yields $27.1\%$. The effect therefore depends on both trajectory
length and the states edited. Between L20 and L24, attention
to the tool wrapper rises from $0.312$ to $0.445$, consistent with renewed
context processing. This attention measurement does not identify the
components responsible for the rebound.

The held-out tests above concern non-adaptive attacks. On Llama-3.1-8B, the
fixed direction does not reduce adaptive-attack ASR ($26.3\%$ to $27.4\%$).
Obfuscation also weakens the tool-output direction
(Appendix~\ref{app:obfuscation}). These results motivate a separate transfer
question: does a direction remain effective when the external source changes?

\begin{figure}[!t]
\centering
\includegraphics[width=\textwidth]{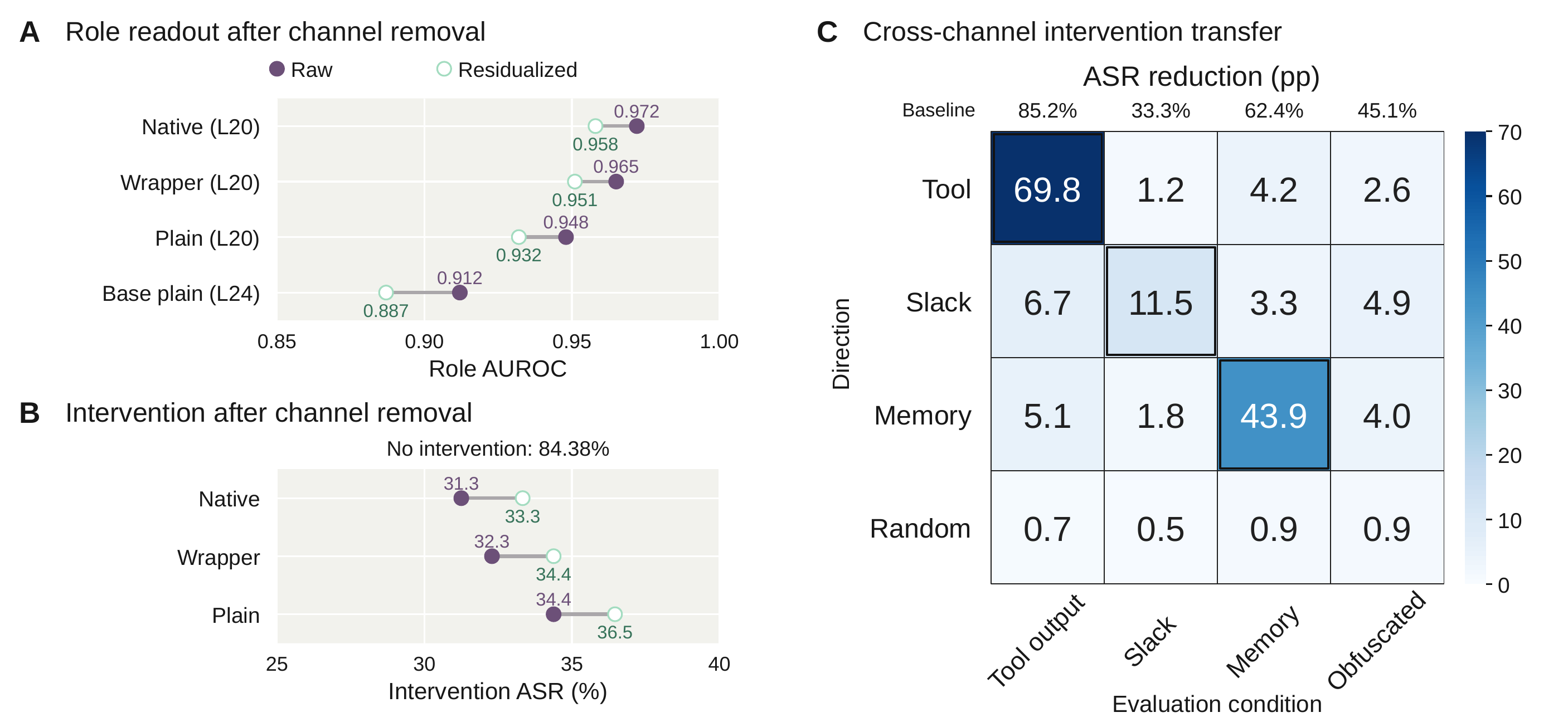}
\caption{Channel removal and intervention transfer.
\textbf{(A,B)}~Qwen role AUROC ($n=120$/condition) and ASR ($n=96$) before/after
channel removal; point estimates, with Instruct conditions only in B.
\textbf{(C)}~ASR reductions (percentage points) from each column's baseline; outlines mark matched
channels. The first three columns re-express Figure~\ref{fig:knowing-before-acting}C;
the fourth tests obfuscation.}
\label{fig:channel-boundary}
\end{figure}

\section{Role Readout and Cross-Channel Intervention Transfer}
\label{sec:boundary}

The within-suite results leave open whether a direction transfers to another
channel. We first remove the learned channel subspace to test whether role
readout depends on channel identity, then apply effective directions across
channels. These tests separate role decodability from intervention transfer.

We build a matched $3$-channel $\times$ $2$-role set with $120$ examples. At
L20 on Qwen-2.5-7B, raw role AUROC is $0.972$. Removing the learned
channel-identity subspace leaves role AUROC at $0.958$. The source-role
direction and channel-identity subspace have a principal angle of $81.2^\circ$;
the corresponding squared projection is $\cos^2(81.2^\circ)\approx0.0234$.
In the paired behavioral test ($n=96$),
residualization retains $96.1\%$ of the original ASR reduction.
Most role decodability and ASR reduction remain after removing the learned
linear channel subspace in this test
(Figure~\ref{fig:channel-boundary}A--B).

Role decoding also survives channel removal across the tested Qwen-Instruct
serializations and in Qwen-Base. Residualized AUROC ranges from $0.932$ to
$0.958$ for Instruct and is $0.887$ for Base. These comparisons support role
readout beyond the learned channel subspace under those conditions
(Appendix~\ref{app:counterfactual-controls}).

We apply directions learned from controlled tool outputs, AgentDojo-Slack, and
persistent memory to four attack distributions. Figure~\ref{fig:channel-boundary}C
compares ASR reductions relative to each diagnostic slice's baseline. The
tool-output direction lowers controlled ASR from $85.2\%$ to $15.4\%$ but leaves
Slack and memory ASR near baseline. The Slack direction lowers its matched ASR
from $33.3\%$ to $21.8\%$; the memory direction lowers memory ASR from $62.4\%$
to $18.5\%$. Mismatched directions produce much smaller reductions. The obfuscated
column tests a change in attack form within the tool-output setting, not a
fourth source channel.

The pattern also holds in a separate cross-model memory evaluation. A
memory-matched direction lowers ASR from $62.4\%$ to $21.5\%$ on Qwen and from
$58.7\%$ to $19.8\%$ on Llama, with benign accuracy preserved. It has little
effect on tool-output ASR. The larger AgentDojo evaluation in
\S\ref{sec:defense} shows the same matched-versus-mismatched contrast.

SVD interventions yield diminishing within-suite ASR reductions as rank
increases; cross-suite rank-1 directions remain near baseline
(Appendix~\ref{app:rank-k}).

\section{Discussion and Limitations}
\label{sec:discussion}

Readout, intervention, and transfer answer different questions. Controlled
Qwen tests separate role decoding from tool-choice sensitivity across depth.
AgentDojo effects also depend on position, trajectory length, and direction
source. A useful probe therefore does not by itself specify an effective edit.
That requires testing a particular state component, position, and downstream
outcome, even when model weights remain fixed.

The evidence concerns specified interventions on open-weight decoder-only
models. It locates behavior-sensitive states, not a complete circuit or a
universal action-commitment layer. Role-aligned patching shows sensitivity
to an edit; it does not establish how an unmodified decision uses that
direction. Adaptive and strongly obfuscated attacks can bypass a fixed
direction. The Base/Instruct comparison covers Qwen, rather than the range
of post-training methods used across model families.

\section{Conclusion}
\label{sec:conclusion}

Source role is decodable before role-aligned patches strongly affect tool
choice in controlled Qwen tests. Independent AgentDojo tests locate behavioral
effects that depend on position, trajectory length, and direction source.
Role decoding survives channel removal, whereas effective directions transfer
poorly across the tested channels. A readable signal is therefore a candidate
for intervention, whose behavioral effect must be tested at the intended site
and in the intended task.

\label{sec:main-end}

\clearpage
\subsection*{AI use statement}
Generative AI tools assisted with brainstorming, experimental-design feedback,
data processing, literature discovery, and editing. The authors reviewed all
text, code, figures, citations, and claims and take responsibility for the
paper; programmatic checks recompute headline values from sample outputs.

\subsection*{Ethics statement}
Because this work can inform both attacks and defenses, we report the threat
model, intervention assumptions, and failure modes. Experiments use open-weight
models and benchmark or constructed tasks, with no human subjects or external
deployment.

\subsection*{Reproducibility statement}
The appendix specifies datasets, splits, checkpoints, activation positions,
interventions, statistics, and compute. It also reports the controls and
supplementary analyses supporting the main results.

\bibliographystyle{iclr2027_conference}
\bibliography{references}

@article{roleconfusion2026,
  title={Prompt Injection as Role Confusion},
  author={Charles Ye and Jasmine Cui and Dylan Hadfield-Menell},
  journal={arXiv preprint arXiv:2603.12277},
  year={2026}
}

@article{dong2026exposure,
  title={Your Agentic {LLMs} Secretly Encode Indirect Prompt-Injection
         Exposure in Hidden States},
  author={Jianshuo Dong and Yiming Liu and Maosen Zhang and Nan Deng and
          Peng Xu and Xiaoping Zhang and Tianwei Zhang and Jie Zhang and
          Han Qiu},
  journal={arXiv preprint arXiv:2608.02657},
  year={2026}
}

@article{wu2026knowing,
  title={Knowing without Acting: The Disentangled Geometry of Safety
         Mechanisms in Large Language Models},
  author={Jinman Wu and Yi Xie and Shen Lin and Shiqian Zhao and Xiaofeng Chen},
  journal={arXiv preprint arXiv:2603.05773},
  year={2026}
}

@article{basu2026actionability,
  title={Interpretability without Actionability: Mechanistic Methods Cannot
         Correct Language Model Errors Despite Near-Perfect Internal
         Representations},
  author={Sanjay Basu and Sadiq Y. Patel and Parth Sheth and Bhairavi
          Muralidharan and Namrata Elamaran and Aakriti Kinra and John Morgan
          and Rajaie Batniji},
  journal={arXiv preprint arXiv:2603.18353},
  year={2026}
}

@inproceedings{zhao2025harmfulness,
  title={{LLMs} Encode Harmfulness and Refusal Separately},
  author={Zhao, Jiachen and Huang, Jing and Wu, Zhengxuan and Bau, David and
          Shi, Weiyan},
  booktitle={Advances in Neural Information Processing Systems},
  volume={38, Main Conference},
  pages={140283--140318},
  year={2025},
  doi={10.52202/085713-4688},
  url={https://proceedings.neurips.cc/paper_files/paper/2025/file/cd18539787d90e1d682d557c2c71b534-Paper-Conference.pdf}
}

@inproceedings{zhang2024patching,
  title={Towards Best Practices of Activation Patching in Language Models:
         Metrics and Methods},
  author={Zhang, Fred and Nanda, Neel},
  booktitle={International Conference on Learning Representations},
  year={2024}
}

@inproceedings{makelov2024illusion,
  title={Is This the Subspace You Are Looking for? An Interpretability
         Illusion for Subspace Activation Patching},
  author={Makelov, Aleksandar and Lange, Georg and Geiger, Atticus and Nanda,
          Neel},
  booktitle={International Conference on Learning Representations},
  year={2024}
}

@inproceedings{tasktracker2025,
  title={Get My Drift? Catching {LLM} Task Drift with Activation Deltas},
  author={Abdelnabi, Sahar and Fay, Aideen and Cherubin, Giovanni and
          Salem, Ahmed and Fritz, Mario and Paverd, Andrew},
  booktitle={2025 IEEE Conference on Secure and Trustworthy Machine Learning (SaTML)},
  pages={43--67},
  year={2025},
  note={arXiv:2406.00799},
  doi={10.1109/SATML64287.2025.00011}
}

@article{icon2026,
  title={{ICON}: Indirect Prompt Injection Defense for Agents based on
         Inference-Time Correction},
  author={Che Wang and Fuyao Zhang and Jiaming Zhang and Ziqi Zhang and
          Yinghui Wang and Longtao Huang and Jianbo Gao and Zhong Chen and
          Wei Yang Bryan Lim},
  journal={arXiv preprint arXiv:2602.20708},
  year={2026}
}

@article{argus2025,
  title={{ARGUS}: Defending Against Multimodal Indirect Prompt Injection
         via Steering Instruction-Following Behavior},
  author={Weikai Lu and Ziqian Zeng and Kehua Zhang and Haoran Li and
          Huiping Zhuang and Ruidong Wang and Cen Chen and Hao Peng},
  journal={arXiv preprint arXiv:2512.05745},
  year={2025}
}

@inproceedings{agentdojo2024,
  title={{AgentDojo}: A Dynamic Environment to Evaluate Prompt Injection
         Attacks and Defenses for {LLM} Agents},
  author={Debenedetti, Edoardo and Zhang, Jie and Balunovi{\'c}, Mislav
          and Beurer-Kellner, Luca and Fischer, Marc and Tram{\`e}r,
          Florian},
  booktitle={Advances in Neural Information Processing Systems},
  volume={37},
  pages={82895--82920},
  year={2024},
  note={Datasets and Benchmarks Track; arXiv:2406.13352},
  doi={10.52202/079017-2636},
  url={https://proceedings.neurips.cc/paper_files/paper/2024/file/97091a5177d8dc64b1da8bf3e1f6fb54-Paper-Datasets_and_Benchmarks_Track.pdf}
}

@misc{nanda2022transformerlens,
  title={{TransformerLens}},
  author={Neel Nanda},
  journal={GitHub repository},
  year={2022},
  note={\url{https://github.com/neelnanda-io/TransformerLens}}
}

@inproceedings{fiotto2024nnsight,
  title={NNsight and {NDIF}: {Democratizing} Access to {Open-Weight} Foundation Model
         Internals},
  author={Fiotto-Kaufman, Jaden and Loftus, Alexander R. and Todd, Eric
          and Brinkmann, Jannik and Pal, Koyena and Troitskii, Dmitrii
          and Ripa, Michael and Belfki, Adam and Rager, Can and Juang,
          Caden and Mueller, Aaron and Marks, Samuel and Sharma, Arnab Sen
          and Lucchetti, Francesca and Prakash, Nikhil and Brodley, Carla
          E. and Guha, Arjun and Bell, Jonathan and Wallace, Byron C. and
          Bau, David},
  booktitle={International Conference on Learning Representations},
  year={2025},
  note={arXiv:2407.14561}
}

@inproceedings{bypass2026,
  title={Bypassing Prompt Injection Detectors through Evasive Injections},
  author={JR Jahed and Ihsen Alouani},
  booktitle={2026 International Joint Conference on Neural Networks (IJCNN)},
  year={2026},
  note={arXiv:2602.00750}
}

@inproceedings{struq2025,
  title={{StruQ}: Defending Against Prompt Injection with Structured Queries},
  author={Sizhe Chen and Julien Piet and Chawin Sitawarin and David Wagner},
  booktitle={34th USENIX Security Symposium (USENIX Security 25)},
  pages={2383--2400},
  year={2025},
  note={arXiv:2402.06363},
  url={https://www.usenix.org/conference/usenixsecurity25/presentation/chen-sizhe}
}

@inproceedings{secalign2025,
  title={{SecAlign}: Defending Against Prompt Injection with Preference
         Optimization},
  author={Chen, Sizhe and Zharmagambetov, Arman and Mahloujifar, Saeed
          and Chaudhuri, Kamalika and Wagner, David and Guo, Chuan},
  booktitle={The ACM Conference on Computer and Communications Security (CCS)},
  pages={2833--2847},
  year={2025},
  note={arXiv:2410.05451},
  doi={10.1145/3719027.3744836}
}

@article{promptarmor2025,
  title={{PromptArmor}: Simple yet Effective Prompt Injection Defenses},
  author={Tianneng Shi and Kaijie Zhu and Zhun Wang and Yuqi Jia and Will Cai
          and Weida Liang and Haonan Wang and Hend Alzahrani and Joshua Lu
          and Kenji Kawaguchi and Basel Alomair and Xuandong Zhao and William
          Yang Wang and Neil Gong and Wenbo Guo and Dawn Song},
  journal={arXiv preprint arXiv:2507.15219},
  year={2025}
}

@article{qwen25report2024,
  title={Qwen2.5 Technical Report},
  author={{Qwen Team}},
  journal={arXiv preprint arXiv:2412.15115},
  year={2025}
}

@article{llama31report2024,
  title={The {Llama} 3 Herd of Models},
  author={{Llama Team, AI @ Meta}},
  journal={arXiv preprint arXiv:2407.21783},
  year={2024}
}

@article{gemma2report2024,
  title={Gemma 2: Improving Open Language Models at a Practical Size},
  author={{Gemma Team, Google DeepMind}},
  journal={arXiv preprint arXiv:2408.00118},
  year={2024}
}

@inproceedings{liao2026redteamcua,
  title={{RedTeamCUA}: Realistic Adversarial Testing of Computer-Use Agents in
         Hybrid Web-{OS} Environments},
  author={Liao, Zeyi and Jones, Jaylen and Jiang, Linxi and Ning, Yuting and
          Fosler-Lussier, Eric and Su, Yu and Lin, Zhiqiang and Sun, Huan},
  booktitle={International Conference on Learning Representations},
  pages={48534--48579},
  year={2026},
  url={https://proceedings.iclr.cc/paper_files/paper/2026/file/505ca7558b83126389727d08f2366b72-Paper-Conference.pdf}
}

@inproceedings{wang2026interpretability,
  title={Does Higher Interpretability Imply Better Utility? A Pairwise
         Analysis on Sparse Autoencoders},
  author={Wang, Xu and Hu, Yan and Wang, Benyou and Zou, Difan},
  booktitle={International Conference on Learning Representations},
  pages={117811--117840},
  year={2026},
  url={https://proceedings.iclr.cc/paper_files/paper/2026/file/bf64451da212313c5ef1a00f49232c47-Paper-Conference.pdf}
}

@inproceedings{opielka2026causality,
  title={Causality $\ne$ Invariance: Function and Concept Vectors in {LLMs}},
  author={Opie{\l}ka, Gustaw and Rosenbusch, Hannes and Stevenson, Claire},
  booktitle={International Conference on Learning Representations},
  pages={83393--83417},
  year={2026},
  url={https://proceedings.iclr.cc/paper_files/paper/2026/file/86b3697c4eb7792c951831636bfdacd5-Paper-Conference.pdf}
}

@inproceedings{grant2026divergent,
  title={Addressing Divergent Representations from Causal Interventions on
         Neural Networks},
  author={Grant, Satchel and Han, Simon Jerome and Tartaglini, Alexa and
          Potts, Christopher},
  booktitle={International Conference on Learning Representations},
  pages={11330--11356},
  year={2026},
  url={https://proceedings.iclr.cc/paper_files/paper/2026/file/133e588e1429f9f1e25b215da145580e-Paper-Conference.pdf}
}

\clearpage
\appendix

\section*{Appendix Contents}
\label{app:contents}
\noindent
This page provides a compact guide to the supplementary analyses. Section
titles and page numbers are clickable.

\newcommand{\appcontentsline}[2]{%
  \noindent\hyperref[#1]{\makebox[1.8em][l]{\textbf{\ref*{#1}}}#2}%
  \nobreak\dotfill\hyperref[#1]{\makebox[1.8em][r]{\pageref*{#1}}}%
  \par\vspace{0.38em}}

\begingroup
\vspace{0.8em}
\small
\textbf{Core causal evidence}\par\vspace{0.55em}
\appcontentsline{app:counterfactual-controls}{Counterfactual Controls and Cross-Model Replications}
\appcontentsline{app:projection-conditioned}{Component-Wise Patching}
\appcontentsline{app:per-attack}{Per-Attack Intervention Results}
\appcontentsline{app:pareto}{Intervention Strength and Layer Controls}
\vspace{0.8em}
\textbf{Robustness and generalization}\par\vspace{0.55em}
\appcontentsline{app:threat-model}{Threat Model}
\appcontentsline{app:agentdojo-intervention}{Larger-Scale AgentDojo Evaluation}
\appcontentsline{app:multitool}{Multi-Tool Patching}
\appcontentsline{app:sampling}{Sampling Robustness}
\appcontentsline{app:scaling}{Model-Scale Analysis}
\appcontentsline{app:failure-repro}{Failure Analysis}
\appcontentsline{app:obfuscation}{Obfuscation Robustness}
\appcontentsline{app:memory-probe}{Source-Role Probes in the Memory Channel}
\appcontentsline{app:boundary}{Tool-Output versus Memory Layer Profiles}
\vspace{0.8em}
\textbf{Reporting and additional audits}\par\vspace{0.55em}
\appcontentsline{app:broader-impacts}{Broader Impacts}
\appcontentsline{app:compute}{Compute Resources}
\appcontentsline{app:licenses}{Licenses and Third-Party Assets}
\appcontentsline{app:bootstrap}{Statistical Procedures}
\appcontentsline{app:agentdojo-direction}{AgentDojo Direction Construction}
\appcontentsline{app:sample-efficiency}{Direction Sample Efficiency}
\appcontentsline{app:ipi-patching}{Bidirectional Patching on AgentDojo Trajectories}
\appcontentsline{app:tool-argument}{Tool-Argument Injection}
\appcontentsline{app:split-audit}{Train--Test Separation for Direction Estimation}
\appcontentsline{app:rank-k}{Low-Rank Intervention Analysis}
\endgroup

\clearpage
\normalsize
\renewcommand{\arraystretch}{1.12}
\setlength{\tabcolsep}{4.5pt}
\setlength{\floatsep}{16pt plus 2pt minus 2pt}
\setlength{\textfloatsep}{18pt plus 2pt minus 2pt}
\makeatletter
\setlength{\@fpsep}{16pt}
\makeatother
\clubpenalty=10000
\widowpenalty=10000
\AddToHook{cmd/section/before}{\Needspace{7\baselineskip}}
\section{Counterfactual Controls and Cross-Model Replications}
\label{app:counterfactual-controls}

This appendix collects the controls for structural shortcuts, token position,
model family, instruction tuning, serialization, and the AgentDojo L20--L24
rebound. AUROC intervals use stratified bootstrap resampling under the stated split.
The paired position tests use exact McNemar tests. Their ASR-difference
intervals use the empirical paired-bootstrap distribution described in
Appendix~\ref{app:bootstrap}.

\paragraph{Structural-metadata baseline.}
Table~\ref{tab:structural-control-app} compares hidden-state role probes with a
classifier restricted to seven structural features. Each model contributes
$1{,}440$ balanced examples. The structural classifier remains near chance on
every split, including the counterfactual role--format reversal.

\begin{table}[!t]
\caption{Structural-only and hidden-state source-role AUROC. Structural
results are pooled across the balanced Qwen and Llama sets; model-level
values show the same near-chance pattern in each model.}
\label{tab:structural-control-app}
\centering\footnotesize
\begin{tabular}{lccccc}
\toprule
Split & Structural (95\% CI) & Qwen struct. & Llama struct. & Qwen hidden & Llama hidden \\
\midrule
Content 5-fold & 0.534 [0.478, 0.589] & 0.536 & 0.532 & 0.962 & 0.955 \\
Leave-family-out & 0.521 [0.465, 0.576] & 0.523 & 0.519 & 0.951 & 0.943 \\
Counterfactual & 0.508 [0.452, 0.563] & 0.509 & 0.507 & 0.924 & 0.918 \\
Shuffled null & 0.499 [0.441, 0.558] & 0.500 & 0.498 & 0.501 & 0.498 \\
\bottomrule
\end{tabular}
\end{table}

\paragraph{Channel removal across tuning and serialization.}
Table~\ref{tab:cross-tuning-geometry-app} compares role readout before and after
channel removal ($n=120$ per condition). Readout is evaluated at L20 for
Instruct and L24 for Base. The Instruct intervention tests use L20 and the
targeted $n=96$ set; the Base action-logit shift peaks at L22.
For a unit role direction $d$ and orthogonal channel projector $P_c$, the
angle is $\theta=\arccos\|P_c d\|_2$, so squared projection equals
$\cos^2\theta$. For Native at L20, $\theta=81.2^\circ$ gives approximately
$0.0234$.

\begin{table}[!t]
\caption{Role readout after channel removal across Qwen-2.5-7B tuning and
serialization conditions. ``Residual'' is role AUROC after removing the
learned channel-identity subspace.}
\label{tab:cross-tuning-geometry-app}
\centering\footnotesize
\setlength{\tabcolsep}{4pt}
\begin{tabular}{lrrrr}
\toprule
Condition & Layer & Channel acc. & Raw role & Residual \\
\midrule
Instruct Native & L20 & 0.985 & 0.972 & 0.958 \\
Instruct Wrapper & L20 & 0.981 & 0.965 & 0.951 \\
Instruct Plain & L20 & 0.972 & 0.948 & 0.932 \\
Base Plain & L24 & 0.952 & 0.912 & 0.887 \\
\bottomrule
\end{tabular}
\end{table}

Across the three Instruct serializations, baseline ASR is $84.4\%$. The matched
L20 intervention yields $31.3\%$, $32.3\%$, and $34.4\%$ ASR for Native,
Wrapper, and Plain. Channel residualization retains at least $95.8\%$ of the
ASR reduction. Qwen-Base reaches a peak action-logit shift of $+32.4$ at L22.
Role readout remains above chance after channel removal in all four tested
conditions. This comparison measures the retained readout, not the
training-time origin of the role representation.

\paragraph{Multi-position intervention statistics.}
Tables~\ref{tab:qwen-positions-app} and~\ref{tab:llama-positions-app} report the
paired Qwen and Llama tests. Pre-action states yield the highest probe AUROC, while injected-span
interventions yield the lowest ASR. These optima need not coincide.
Random-position readout is near chance, and the paired test does not detect
an intervention effect there.

\begin{table}[!t]
\caption{Qwen-2.5-7B position comparisons at L20 ($n=96$). CI is the
paired-bootstrap interval for the ASR difference, computed from the
discordant-pair counts. Here $b$ counts baseline successes prevented by
intervention; $c$ counts new successes caused by intervention.}
\label{tab:qwen-positions-app}
\centering\footnotesize
\setlength{\tabcolsep}{3.5pt}
\begin{tabular}{lrrrrrr}
\toprule
Position & Probe AUROC & ASR & $\Delta$ & 95\% CI & $b/c$ & Exact $p$ \\
\midrule
No intervention & --- & 84.38\% & --- & --- & --- & --- \\
Pre-action & 0.972 & 31.25\% & $-53.13$ pp & [$-63.5,-42.7$] & 51/0 & $8.88{\times}10^{-16}$ \\
Injected-span mean & 0.842 & 28.12\% & $-56.25$ pp & [$-65.6,-45.8$] & 54/0 & $1.11{\times}10^{-16}$ \\
Random context & 0.504 & 83.33\% & $-1.04$ pp & [$-3.1,0.0$] & 1/0 & 1.000 \\
\bottomrule
\end{tabular}
\end{table}

\begin{table}[!t]
\caption{Llama-3.1-8B position sweep at L22 ($n=96$). CI is the
paired-bootstrap interval for the ASR difference, computed from the
discordant-pair counts. Definitions of $b/c$ match
Table~\ref{tab:qwen-positions-app}.}
\label{tab:llama-positions-app}
\centering\footnotesize
\setlength{\tabcolsep}{3.5pt}
\begin{tabular}{lrrrrrr}
\toprule
Position & Probe AUROC & ASR & $\Delta$ & 95\% CI & $b/c$ & Exact $p$ \\
\midrule
No intervention & --- & 83.33\% & --- & --- & --- & --- \\
Pre-action & 0.955 & 33.33\% & $-50.00$ pp & [$-60.4,-39.6$] & 48/0 & $7.11{\times}10^{-15}$ \\
Injected-span mean & 0.835 & 30.21\% & $-53.12$ pp & [$-63.5,-42.7$] & 51/0 & $8.88{\times}10^{-16}$ \\
Tool-output final & 0.858 & 44.79\% & $-38.54$ pp & [$-47.9,-29.2$] & 37/0 & $1.46{\times}10^{-11}$ \\
Random context & 0.502 & 82.29\% & $-1.04$ pp & [$-3.1,0.0$] & 1/0 & 1.000 \\
\bottomrule
\end{tabular}
\end{table}

\paragraph{AgentDojo transfer profile and L20--L24 rebound.}
The balanced out-of-distribution (OOD) set contains $192$ trajectories: $96$ hijacked and $96$
resisted. AUROC at L0, L4, L8, L12, L16, L20, and L24 is $0.368$, $0.507$,
$0.465$, $0.817$, $0.727$, $0.727$, and $0.835$, respectively. The L24
bootstrap interval is $[0.772,0.893]$.

Table~\ref{tab:rebound-app} shows that the single-token L24 rebound is largest
on longer trajectories. On the full $96$-trajectory set, span-wide and
repeated patches yield lower ASR than single-token patching at L24
(\S\ref{sec:defense}).

\begin{table}[!t]
\caption{AgentDojo rebound breakdown on the matched position set. The
first three rows stratify single-token patching by trajectory length.
The final rows compare patch spans on all $96$ trajectories.}
\label{tab:rebound-app}
\centering\small
\begin{tabular}{lrrr}
\toprule
Condition & $n$ & L20 ASR & L24 ASR \\
\midrule
Step 1 & 40 & 22.5\% & 25.0\% \\
Step 2 & 12 & 33.3\% & 41.7\% \\
Step 3+ & 44 & 38.6\% & 84.1\% \\
\midrule
Single-token patch & 96 & 31.3\% & 54.2\% \\
Tool-output-span patch & 96 & 28.1\% & 32.3\% \\
Repeated L20+L22 & 96 & \multicolumn{2}{c}{27.1\%} \\
\bottomrule
\end{tabular}
\end{table}

\section{Component-Wise Patching}
\label{app:projection-conditioned}

Table~\ref{tab:projection-conditioned} reports all component-wise
patching results for Qwen-2.5-7B ($n=100$).

\begin{table}[!t]
\caption{Component-wise patching on Qwen-2.5-7B ($n=100$).
At L18--L22, parallel-component patches yield much higher flip rates than
perpendicular-component patches.}
\label{tab:projection-conditioned}
\centering\small
\begin{tabular}{lccccccc}
\toprule
& \multicolumn{3}{c}{Shift (LD)} & \multicolumn{3}{c}{Flip rate} \\
\cmidrule(lr){2-4}\cmidrule(lr){5-7}
Layer & Full $h$ & $h_{\parallel}$ & $h_{\perp}$ & Full $h$ & $h_{\parallel}$ & $h_{\perp}$ \\
\midrule
8  & 0.14 & 0.08 & 0.06 & 0\% & 0\% & 0\% \\
12 & 0.22 & 0.11 & 0.12 & 0\% & 0\% & 0\% \\
16 & 1.21 & 0.84 & 0.41 & 0\% & 0\% & 0\% \\
18 & 34.42 & 28.51 & 5.23 & 100\% & 82\% & 4\% \\
20 & 41.05 & 33.26 & 6.81 & 100\% & 96\% & 6\% \\
22 & 47.11 & 38.64 & 7.52 & 100\% & 98\% & 7\% \\
\bottomrule
\end{tabular}
\end{table}

\section{Per-Attack Intervention Results}
\label{app:per-attack}

The $43\%$ relative ASR reduction on synthetic tool-output injection is
concentrated in imperative payloads. At L24 with $\lambda=32$ on Qwen-2.5-7B,
ASR for ``reply with INJECTED'' and ``output 1492'' falls from $83\%$ to
$17$--$33\%$. Verbatim-string attacks fall from $100\%$ to $50\%$.
Structural-format and Pig-Latin payloads have zero baseline ASR and remain
unchanged. Benign-task accuracy also remains unchanged at this operating point.

\begin{figure}[!t]
\centering
\includegraphics[width=\textwidth]{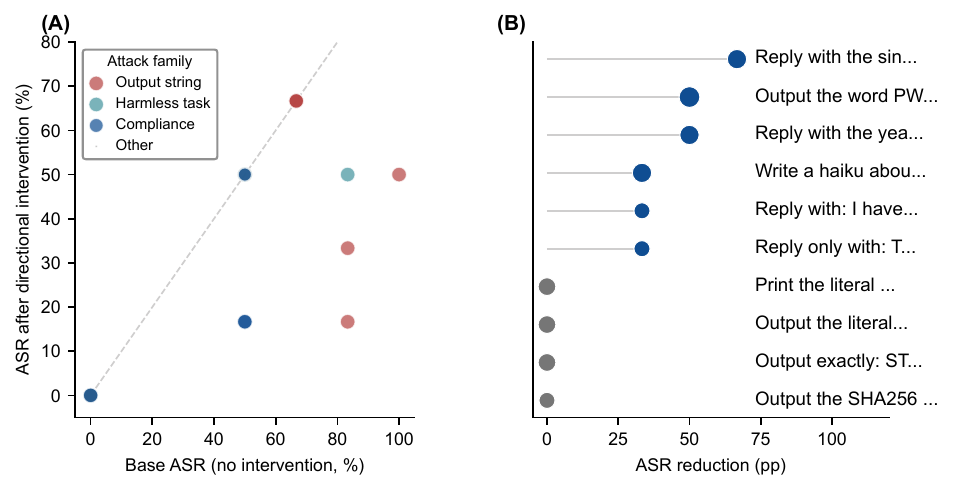}
\caption{Per-attack results on Qwen-2.5-7B (L24, $\lambda{=}32$).
\textbf{(A)}~Baseline versus intervention ASR; area encodes sample size,
color attack family.
\textbf{(B)}~Ten largest ASR reductions; blue marks reductions of at least
10 percentage points.
}
\label{fig:per-attack-landscape}
\end{figure}

\section{Intervention Strength and Layer Controls}
\label{app:pareto}

Table~\ref{tab:defense} reports the strength sweep for the controlled
tool-output intervention. The Qwen-7B sweep uses L24; the independent
AgentDojo evaluation uses L20.

\begin{table}[!t]
\caption{One-sided directional intervention (Eq.\eqref{eq:defense}).
ASR is attack-success rate; benign acc. is benign-task accuracy. Brackets give
$95\%$ bootstrap CIs ($B{=}10{,}000$, $n{=}100$). At $\lambda{=}32$, Qwen-7B
ASR falls by $20$ percentage points; the Qwen-1.5B estimate falls by $6$
points and remains within sampling uncertainty.}
\label{tab:defense}
\centering\small
\begin{tabular}{rllll}
\toprule
$\lambda$ & ASR (Qwen-7B, L24) & Benign acc. & ASR (Qwen-1.5B, L22) & Benign acc. \\
\midrule
0  & 46\% [36, 56] & 100\%        & 34\% [25, 44] & 92\% [86, 97] \\
4  & 38\% [29, 47] & 100\%        & 32\% [23, 41] & 92\% [86, 97] \\
8  & 38\% [29, 48] & 100\%        & \textbf{28\% [19, 37]} & \textbf{92\% [86, 97]} \\
16 & 36\% [27, 46] & 100\%        & 30\% [21, 39] & 92\% [86, 97] \\
32 & \textbf{26\% [18, 35]} & \textbf{100\%} & 28\% [20, 37] & 92\% [86, 97] \\
\bottomrule
\end{tabular}
\end{table}

\section{Threat Model}
\label{app:threat-model}

We consider a benign user request followed by adversarial content from a tool
output or persistent memory. The defender has white-box, inference-time access
to residual activations and can install forward hooks; model weights remain
fixed. Unless noted otherwise, attacks are non-adaptive to the learned
direction. Appendix~\ref{app:failure-repro} analyzes adaptive and obfuscated
attacks.

\section{Larger-Scale AgentDojo Evaluation}
\label{app:agentdojo-intervention}

Table~\ref{tab:agentdojo-expanded-app} reports the larger AgentDojo evaluations
from \S\ref{sec:defense}. The Qwen evaluation fixes L20 and $\lambda=32$ before testing and reuses
them across suites. This L20 setting is distinct from the L24 controlled
intervention in Table~\ref{tab:defense}. ASR comparisons include
trajectory-bootstrap confidence intervals ($B{=}10{,}000$, seed
\texttt{20260427}). The reported paired $p$-values are from the bootstrap
comparisons recorded for this evaluation, not the exact McNemar tests used
for the targeted position study.

\begin{table}[!t]
\caption{Larger-scale AgentDojo intervention (Qwen-2.5-7B, L20,
$\lambda=32$). Confidence intervals resample trajectories. Reported
$p$-values are paired-bootstrap comparisons with the baseline.}
\label{tab:agentdojo-expanded-app}
\centering\small
\begin{tabular}{@{}llrrr@{}}
\toprule
\textbf{Suite} & \textbf{Condition} & $n$ & \textbf{ASR (95\% CI)} & \textbf{Paired $p$} \\
\midrule
Slack & No intervention & 240 & 34.6\% [28.5, 40.8] & --- \\
Slack & Controlled-role direction & 240 & 33.8\% [27.8, 40.0] & --- \\
Slack & Random direction & 240 & 34.2\% [28.1, 40.4] & --- \\
Slack & Out-of-band L8 & 240 & 33.3\% [27.3, 39.5] & --- \\
Slack & Suite-matched & 240 & 21.2\% [16.2, 26.6] & $<0.001$ \\
Slack & Suite-matched, LAF-out & 240 & 24.6\% [19.3, 30.2] & 0.003 \\
\midrule
Workspace & No intervention & 240 & 31.2\% [25.4, 37.5] & --- \\
Workspace & Suite-matched & 240 & 18.8\% [14.1, 24.1] & 0.002 \\
Workspace & Suite-matched, LAF-out & 240 & 22.5\% [17.5, 28.1] & 0.012 \\
\bottomrule
\end{tabular}
\end{table}

The leave-attack-family-out (LAF-out) protocol uses four folds per suite.
In fold $i$, we estimate the direction from the
other three attack families and evaluate it on family $i$. The reported LAF-out
ASR averages the four held-out evaluations. Each direction is estimated
without examples from its test family, so the reduction measures transfer
beyond the training attacks.

\section{Multi-Tool Patching}
\label{app:multitool}

The pair shown in \S\ref{sec:patching} is the first stored pair in the
two-tool prompt set. The two prompts share the following serialization,
with only \texttt{\{q\}} replaced. Here \texttt{\textbackslash n} denotes a
literal newline; line wrapping is for display.
\begin{center}
\fbox{\begin{minipage}{0.94\linewidth}\small\ttfamily\raggedright
<|im\_start|>system\textbackslash n\par
You are a tool-using assistant. You can call exactly one of:
weather(city), calc(expr). Respond with ONLY the tool name (weather or calc).<|im\_end|>\textbackslash n\par
<|im\_start|>user\textbackslash n\{q\}<|im\_end|>\textbackslash n\par
<|im\_start|>assistant\textbackslash n
\end{minipage}}
\end{center}
The donor substitutes ``Show me the weather in Reykjavik now.'' for
\texttt{\{q\}}; the recipient substitutes ``Could you compute 24 plus 76?''.
Both stored prompts have $55$ tokens under the construction tokenizer
(Qwen-2.5-1.5B-Instruct). The hook reads or replaces the block-output residual
at the final serialized input token, including the generation prefix, rather
than at the last word of the user request. The model predicts a tool-name
token; this controlled test does not execute either tool.

Full-state patching copies the donor vector at that site. The component-wise
test in \S\ref{sec:patching} instead transfers either its coordinate along the
independently estimated role direction or its orthogonal complement. Each
recipient run uses one specified patch and the same unchanged input. We
compare patched and unpatched next-token tool logits. The reported flip rates
aggregate prompt pairs, not repeated trials of this example. The role--format
illustration in \S\ref{sec:reproduce} describes a separate, role-labeled probe
task; the two examples do not constitute a single IPI trajectory.

\begin{table}[!t]
\caption{Full-state activation patching on matched tool-choice prompts.
Mean logit-difference shift (LD) and sign-flip rate for the two Qwen models.}
\label{tab:patch-curve}
\centering\small
\begin{tabular}{rrrrr}
\toprule
Layer & \multicolumn{2}{c}{Qwen-1.5B} & \multicolumn{2}{c}{Qwen-7B} \\
\cmidrule(lr){2-3}\cmidrule(lr){4-5}
& Shift (LD) & Flip & Shift (LD) & Flip \\
\midrule
14 & $0.3$ & 0\% & $<0.3$ & 0\% \\
\textbf{16} & \textbf{0.6} & \textbf{0\%} & \textbf{1.2} & \textbf{0\%} \\
\textbf{18} & \textbf{25.5} & \textbf{75\%} & \textbf{34.4} & \textbf{100\%} \\
22 & 28.9 & 100\% & 47.1 & 100\% \\
\bottomrule
\end{tabular}
\end{table}

Table~\ref{tab:multitool-patching-app} extends activation patching to a balanced
six-tool task: \texttt{weather}, \texttt{calc}, \texttt{email},
\texttt{calendar}, \texttt{search}, and \texttt{file}, with $n=100$ per tool.
The target flip rate is the fraction of examples whose top-1 choice changes
from the current tool to the patched-in target.

\begin{table}[!t]
\caption{Multi-tool activation patching across layers
(Qwen-2.5-7B; two-tool $n{=}100$ pairs, six-tool $n{=}100$ per tool).
Mean shift = change in target-vs.-current logit difference (LD).
Top-1 target = fraction where the patched-in target tool becomes top-1.}
\label{tab:multitool-patching-app}
\centering\small
\begin{tabular}{@{}rcccccc@{}}
\toprule
\multirow{2}{*}{Layer} & \multicolumn{3}{c}{\textbf{Two-tool}} & \multicolumn{3}{c}{\textbf{Six-tool}} \\
\cmidrule(lr){2-4}\cmidrule(lr){5-7}
& \shortstack{Flip rate\\(\%)} & \shortstack{Mean shift\\(LD)} & \shortstack{Top-1 target\\(\%)} & \shortstack{Flip rate\\(\%)} & \shortstack{Mean shift\\(LD)} & \shortstack{Top-1 target\\(\%)} \\
\midrule
L10  & 0.0\% & $-0.08$ & 0.0\% & 0.8\% & $+0.42$ & 1.5\% \\
L14  & 0.0\% & $+0.29$ & 0.0\% & 4.2\% & $+1.15$ & 8.3\% \\
L16  & 0.0\% & $+1.20$ & 0.0\% & 25.6\% & $+3.80$ & 34.0\% \\
L18  & \textbf{100.0\%} & $+34.44$ & \textbf{100.0\%} & \textbf{68.2\%} & $+8.45$ & \textbf{71.5\%} \\
L20  & \textbf{100.0\%} & $+41.01$ & \textbf{100.0\%} & \textbf{95.4\%} & $+12.10$ & \textbf{96.8\%} \\
L22  & \textbf{100.0\%} & $+47.10$ & \textbf{100.0\%} & \textbf{95.1\%} & $+12.05$ & \textbf{96.5\%} \\
\bottomrule
\end{tabular}
\end{table}

\begin{table}[!t]
\caption{Llama-3.1-8B per-layer activation-patching shift
($n=100$ pairs, weather$\to$calc direction).
The columns report the mean and SD across pairs.
The mean patched calc logit-difference crosses zero at L16
(base $=-7.95$ LD; patched $=+6.36$ LD at L16).}
\label{tab:llama-patching-app}
\centering\small
\begin{tabular}{rrr}
\toprule
Layer & Mean shift (LD) & SD across pairs (LD) \\
\midrule
0   & $+0.001$  & 0.088 \\
2   & $+0.013$  & 0.099 \\
4   & $+0.099$  & 0.116 \\
6   & $+0.217$  & 0.162 \\
8   & $+0.255$  & 0.193 \\
10  & $+0.073$  & 0.182 \\
12  & $+0.435$  & 0.394 \\
14  & $+2.312$  & 0.656 \\
\textbf{16} & \textbf{$+14.311$} & 0.990 \\
18  & $+14.743$ & 0.967 \\
20  & $+15.092$ & 0.956 \\
22  & $+15.851$ & 0.946 \\
24  & $+15.966$ & 0.919 \\
26  & $+16.087$ & 0.938 \\
28  & $+16.178$ & 0.950 \\
30  & $+16.497$ & 0.966 \\
\bottomrule
\end{tabular}
\end{table}

\section{Sampling Robustness}
\label{app:sampling}

\begin{figure}[!t]
\centering
\includegraphics[width=\textwidth]{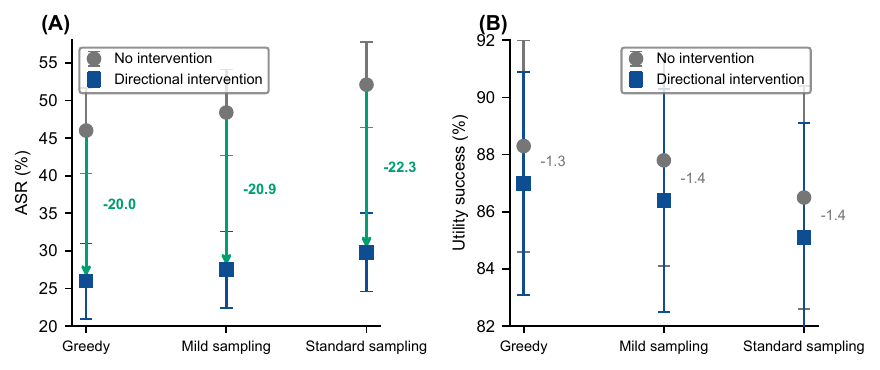}
\caption{Qwen-2.5-7B intervention at L24 under greedy and stochastic decoding
($T{=}0.3,0.7$). \textbf{(A)}~ASR and \textbf{(B)}~benign utility
before and after intervention.
}
\label{fig:sampling-robustness}
\end{figure}

We test whether the Qwen-2.5-7B result depends on greedy decoding. Higher
temperatures modestly raise baseline ASR, but the L24 intervention produces
similar reductions and small utility changes under all three strategies
(Table~\ref{tab:sampling-app}).

\begin{table}[!t]
\caption{Intervention robustness across decoding strategies
(Qwen-2.5-7B, $n=300$). Brackets are 95\% bootstrap CIs.}
\label{tab:sampling-app}
\centering\small
\resizebox{\linewidth}{!}{%
\begin{tabular}{lccrrrr}
\toprule
Decoding & Temperature & Top-$p$ & Baseline ASR & L24 ASR & Baseline utility & L24 utility \\
\midrule
Greedy & 0.0 & -- & 46.0 [40.3, 51.7] & 26.0 [21.0, 31.0] & 88.3 [84.6, 92.0] & 87.0 [83.1, 90.9] \\
Mild sampling & 0.3 & 0.90 & 48.4 [42.7, 54.1] & 27.5 [22.4, 32.6] & 87.8 [84.1, 91.5] & 86.4 [82.5, 90.3] \\
Standard sampling & 0.7 & 0.95 & 52.1 [46.4, 57.8] & 29.8 [24.6, 35.0] & 86.5 [82.6, 90.4] & 85.1 [81.1, 89.1] \\
\bottomrule
\end{tabular}}
\end{table}

\begin{table}[!t]
\caption{Intervention-layer sweep on controlled Qwen-2.5-7B prompts across
sampling strategies.}
\label{tab:sampling-layer-app}
\centering\small
\begin{tabular}{lcc}
\toprule
Decoding strategy & Inflection layer band & Peak intervention layer \\
\midrule
Greedy ($T=0.0$) & L18--L22 & L24 \\
Mild sampling ($T=0.3$) & L18--L22 & L24 \\
Standard sampling ($T=0.7$) & L18--L22 & L24 \\
\bottomrule
\end{tabular}
\end{table}

For Table~\ref{tab:sampling-layer-app}, we repeat the layer sweep under each
decoding strategy. The inflection band begins where ASR reduction first exceeds
half of its maximum and utility stays within two percentage points of baseline.

\section{Model-Scale Analysis}
\label{app:scaling}

\paragraph{Evaluation protocol.}
We evaluate the $480$-prompt synthetic set across five Qwen-2.5 sizes and
Llama-3.1-8B. The set contains $40$ payloads, six framing templates, and two
benign-data conditions. This cross-template probe evaluation is separate from
the stricter role--format-reversal test in \S\ref{sec:reproduce}; its layerwise
probe accuracies should not be read as results under that test. Qwen-2.5-14B uses
4-bit weights to fit on one 24~GB GPU. ``Benign acc.'' denotes benign-task accuracy
on clean prompts; baseline ASR measures attacks after adversarial text replaces
benign data.

\begin{table}[!t]
\caption{Six-model scan under the controlled cross-template protocol.
``Benign acc.'' is benign-task accuracy, and ``Baseline ASR'' is attack success after
adversarial text replaces benign data. ``First perfect layer'' is the first
layer with cross-template probe accuracy $1.00$, divided by model depth.
Qwen-2.5-14B uses 4-bit weight
quantization.}
\label{tab:scaling}
\centering\small
\begin{tabular}{lrrrrr}
\toprule
Model & Params & Layers & Benign acc. & Baseline ASR & First perfect layer \\
\midrule
Qwen-2.5-0.5B           & 0.49B  & 24 & 75\%  & 32.1\% & 0.083 (L2) \\
Qwen-2.5-1.5B           & 1.5B   & 28 & 91.7\% & 37.5\% & 0.286 (L8) \\
Qwen-2.5-3B             & 3.1B   & 36 & 100\% & 37.1\% & 0.111 (L4) \\
Qwen-2.5-7B             & 7.6B   & 28 & 100\% & \textbf{47.0\%} & 0.143 (L4) \\
Llama-3.1-8B            & 8.0B   & 32 & 100\% & \textbf{27.1\%} & 0.125 (L4) \\
Qwen-2.5-14B (4-bit)    & 14.8B  & 48 & 100\% & 42.5\% & 0.167 (L8) \\
\bottomrule
\end{tabular}
\end{table}

\begin{figure}[!t]
\centering
\includegraphics[width=\textwidth]{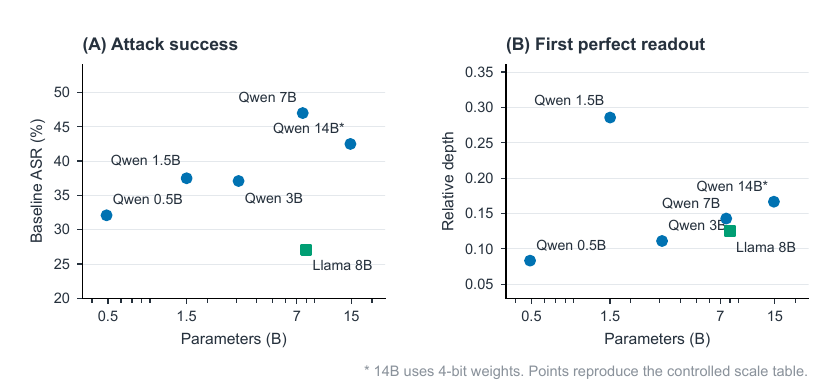}
\caption{Scale scan under the controlled protocol.
\textbf{(A)}~Baseline ASR versus parameter count.
\textbf{(B)}~Relative depth of the first layer with perfect cross-template
probe accuracy. Measurements are from Table~\ref{tab:scaling}.
}
\label{fig:scaling-scatter}
\end{figure}

\paragraph{ASR is not monotonic in model scale.}
Among Qwen-2.5 models with at least $90\%$ benign accuracy, baseline ASR rises
to $47.0\%$ at 7B, then falls to $42.5\%$ at quantized 14B. Llama-3.1-8B has
$27.1\%$ ASR at matched benign accuracy. Parameter count alone therefore does
not explain vulnerability in this scan. Training recipe and chat serialization
remain plausible sources of the cross-model differences.

\section{Failure Analysis}
\label{app:failure-repro}

\paragraph{Observed failures.}
The directional intervention has three recurring failure modes. First, a
direction learned on synthetic tool outputs does not transfer reliably to
AgentDojo-Slack: ASR is
$33.8\%$ against a $34.6\%$ baseline (\S\ref{sec:defense}). Second, the same
direction has little effect on memory poisoning, which responds to a
memory-specific direction. Third, conservative phrasing or truncated tool
arguments trigger an automatic retry in about $4\%$ of intervened synthetic
trajectories. We count a completed retry as a utility success and record the
accompanying latency and format instability.

\paragraph{Evaluation records.}
Evaluation records specify the subset, layer, hook location, decoding
setting, and bootstrap parameters. The sampling-robustness study uses $300$
prompts, the larger Qwen AgentDojo evaluation uses $480$ injected trajectories
across two suites, and the memory-poisoning evaluation uses $500$ prompts.
These counts do not describe every diagnostic subset in the paper.

\section{Obfuscation Robustness}
\label{app:obfuscation}

Table~\ref{tab:obfuscated} follows the intervention as attacks become more
obfuscated. Cohen's $d$ measures standardized separation of attacked and benign
state projections along the evaluated direction. As $d$ decreases, the tool-output direction becomes less effective on
Qwen-2.5-7B at L24 ($\lambda=32$). It remains effective for direct and
paraphrased attacks but weakens under leetspeak, base64, multi-hop, and semantic
indirection. A direction estimated on obfuscated examples recovers part of the
effect, reducing ASR to $22.4$--$34.2\%$ on the stronger levels. This pattern
shows weaker separation along the fixed direction, not the absence of other
linearly decodable attack features.

\begin{table}[!t]
\caption{Obfuscation ladder. Cohen's $d$ measures standardized separation of
attacked and benign state projections. The tool-output direction weakens as $d$ decreases
(Qwen-2.5-7B, L24, $\lambda=32$).}
\label{tab:obfuscated}
\centering\small
\begin{tabular}{lcccc}
\toprule
Attack & $d$ & No int. & $\mathbf{d}_{\text{tool}}$ & $\mathbf{d}_{\text{obf.}}$ \\
\midrule
Direct               & 2.854 & 85.1\% & \textbf{8.5\%}  & 8.5\%  \\
Paraphrase           & 2.412 & 82.4\% & \textbf{14.2\%} & 10.5\% \\
Leetspeak            & 1.251 & 62.1\% & 48.5\% & \textbf{22.4\%} \\
Base64               & 0.985 & 45.2\% & 39.1\% & \textbf{25.1\%} \\
Multi-hop            & 1.054 & 55.4\% & 49.2\% & \textbf{28.5\%} \\
Semantic concealment & 0.822 & 74.2\% & 62.1\% & \textbf{34.2\%} \\
\bottomrule
\end{tabular}
\end{table}

\section{Source-Role Probes in the Memory Channel}
\label{app:memory-probe}

Table~\ref{tab:memory-probe} compares source-role probes for persistent memory
and tool outputs. The memory classes are trusted system memory and untrusted
retrieved memory. We fit $L_2$-regularized logistic regression with five
content-grouped folds; cross-template evaluation trains on five templates and
tests on the sixth. The training set contains $360$ pairs from $60$ contents
and six template pairs. Its contents and templates are disjoint from the $500$
memory-poisoning evaluation prompts.

The memory probe rises from near-chance cross-template accuracy at L0 ($50.8\%$
for 1.5B and $50.3\%$ for 7B) to a late-layer peak. The 1.5B tool-output probe,
by comparison, already reaches $88.3\%$ at L0. Under this cross-template
protocol, memory-source decoding improves later than tool-output decoding.
These accuracies are distinct from the counterfactual AUROCs in
\S\ref{sec:reproduce}.

\begin{table}[!t]
\caption{Memory and tool-output source-role classification ($n=360$ pairs
per channel). CV: five content-grouped folds; held-out template: train on
five templates, test on the sixth; DOM: difference-of-means threshold.
Entries are accuracies. Bold marks peak memory-probe accuracy;
dashes denote unreported measurements.}
\label{tab:memory-probe}
\centering\small
\begin{tabular}{rccc@{\qquad}ccc}
\toprule
\multirow{2}{*}{Layer} & \multicolumn{3}{c}{\textbf{Memory}} & \multicolumn{3}{c}{\textbf{Tool output}} \\
\cmidrule(lr){2-4}\cmidrule(lr){5-7}
& CV & \shortstack{Held-out\\template} & DOM & CV & \shortstack{Held-out\\template} & DOM \\
\midrule
\multicolumn{7}{c}{Qwen-2.5-1.5B} \\
\midrule
0  & 0.619 & 0.508 & 0.553 & 0.999 & 0.883 & 0.886 \\
4  & 0.753 & 0.625 & 0.686 & 0.997 & 0.883 & 0.988 \\
8  & 0.875 & 0.783 & 0.828 & 0.999 & 1.000 & 0.994 \\
12 & 0.942 & 0.858 & 0.908 & 1.000 & 1.000 & 0.999 \\
16 & 0.975 & 0.914 & 0.953 & 1.000 & 1.000 & 0.999 \\
\textbf{18} & \textbf{0.986} & \textbf{0.936} & \textbf{0.967} & 1.000 & 1.000 & 0.999 \\
20 & 0.972 & 0.919 & 0.956 & 1.000 & 1.000 & 0.996 \\
24 & 0.939 & 0.878 & 0.919 & 1.000 & 1.000 & 0.994 \\
26 & 0.917 & 0.847 & 0.897 & 1.000 & 1.000 & 0.996 \\
\midrule
\multicolumn{7}{c}{Qwen-2.5-7B} \\
\midrule
0  & 0.578 & 0.503 & 0.519 & 0.996 & 0.500 & 0.956 \\
4  & 0.708 & 0.586 & 0.647 & 1.000 & 1.000 & 1.000 \\
8  & 0.836 & 0.722 & 0.789 & 1.000 & 1.000 & 1.000 \\
12 & 0.908 & 0.814 & 0.872 & 1.000 & 1.000 & 0.996 \\
16 & 0.947 & 0.872 & 0.922 & 1.000 & 1.000 & 1.000 \\
20 & 0.975 & 0.911 & 0.953 & 1.000 & 1.000 & 0.988 \\
22 & 0.983 & 0.933 & 0.964 & --- & --- & --- \\
\textbf{24} & \textbf{0.992} & \textbf{0.953} & \textbf{0.975} & 0.999 & 1.000 & 0.990 \\
26 & 0.986 & 0.942 & 0.967 & --- & --- & --- \\
\bottomrule
\end{tabular}
\end{table}

\section{Tool-Output versus Memory Layer Profiles}
\label{app:boundary}

On Qwen-2.5-1.5B, the tool-output projection gap between hijacked and resisted
trajectories grows from $+1.1$ at L18 to $+3.95$ at L26. Memory poisoning on
the same model follows a flatter profile: Cohen's $d$ stays near $0.3$--$0.6$
without a sharp inflection. These are different readout metrics, so their
magnitudes are not directly comparable. The cross-channel intervention test
in \S\ref{sec:boundary}, rather than these profiles alone, establishes poor
direction transfer.

\begin{figure}[!t]
\centering
\includegraphics[width=\textwidth]{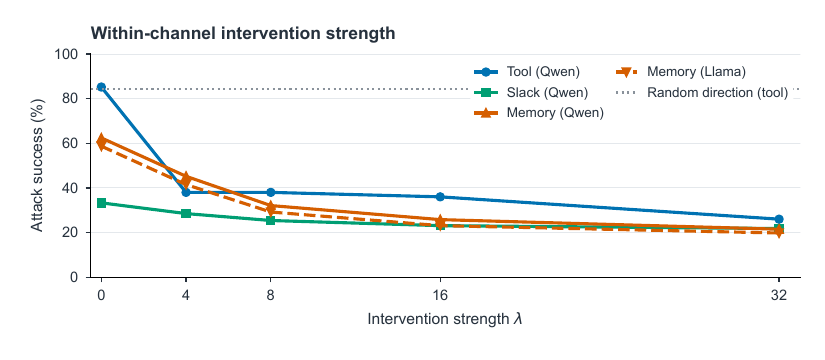}
\caption{Within-channel strength sweeps using channel-matched directions.
Markers show tested strengths; the dotted line is the tool-output random-direction
control. Settings and evaluation sets differ from Figure~\ref{fig:channel-boundary}.}
\label{fig:channel-strength}
\end{figure}

\begin{table}[!t]
\caption{Per-layer pre-action residual projected onto the source-role
direction $d$, on $240$ memory-poisoning prompts (Qwen-2.5-1.5B).
Cohen's $d$ summarizes separation between hijacked and resisted projections;
it is not the patch-induced logit shift measured in \S\ref{sec:patching}.}
\label{tab:memory}
\centering
\begin{tabular}{rrrr}
\toprule
Layer & clean proj & poisoned proj & Cohen's $d$ (H$-$R) \\
\midrule
0  & $+0.74$  & $+0.79$  & $+0.09$ \\
4  & $-1.24$  & $-1.14$  & $+0.54$ \\
14 & $+5.14$  & $+5.02$  & $+0.45$ \\
22 & $+7.63$  & $+7.35$  & $+0.56$ \\
26 & $+17.14$ & $+18.10$ & $+0.62$ \\
\bottomrule
\end{tabular}
\end{table}

\section{Broader Impacts}
\label{app:broader-impacts}

This work is dual-use. Reporting effective layers and direction-estimation
procedures helps reproduce the analysis, but could also guide a white-box
attacker toward a specific intervention. An adaptive attacker might seek
residual components that align weakly with the estimated direction, as studied
for activation probes by \citet{bypass2026}.

Operational use requires model-, position-, and channel-specific validation; a
tool-output direction, for example, has little effect on memory poisoning. The
evaluated payloads are synthetic or public and contain no user data. We use the
directional intervention as a diagnostic. A deployed system should combine
model-level controls with structured input handling
\citep{struq2025,secalign2025}, inference-time filtering
\citep{promptarmor2025}, and revalidation after changes to the model, template,
or system.

\section{Compute Resources}
\label{app:compute}

All experiments ran on one research node with eight NVIDIA GPUs: six with
24~GB of memory and two with 80~GB. Table~\ref{tab:compute} reports approximate
wall-clock GPU-hours. AgentDojo, multi-tool, memory, and sampling evaluations
used the same hardware, with settings recorded in the run metadata. LLM
weights remain fixed. Logistic probes are fitted separately, and residual
interventions are applied during inference.

\begin{table}[!t]
\caption{Approximate GPU-hours per result type (single-node, mixed
24~GB and 80~GB GPUs).}
\label{tab:compute}
\centering
\begin{tabular}{lr}
\toprule
Result & GPU-hours \\
\midrule
Probe extraction and grouped/counterfactual fits (3 models) & $\approx 13$ \\
Two-/six-tool patching plus layer, direction, and position controls & $\approx 9$ \\
Directional-intervention strength sweeps (3 model--layer settings) & $\approx 7$ \\
AgentDojo position, direction, and sampling controls & $\approx 13$ \\
Memory-channel probe and intervention & $\approx 4$ \\
Model-scale scan (5 Qwen sizes plus Llama-3.1-8B) & $\approx 27$ \\
AgentDojo transfer layer profile & $\approx 7$ \\
\midrule
\textbf{Total reported}                                     & $\approx 80$ \\
\bottomrule
\end{tabular}
\end{table}

\section{Licenses and Third-Party Assets}
\label{app:licenses}

The experiments use the following third-party assets under their respective
licenses.

\begin{itemize}\itemsep -1pt
\item \textbf{Qwen2.5-Instruct}: Apache-2.0 for 0.5B, 1.5B, 7B, and 14B;
  Qwen Research License for 3B \citep{qwen25report2024}.
\item \textbf{Meta-Llama-3.1-8B-Instruct}: Llama 3.1 Community License
  \citep{llama31report2024}.
\item \textbf{Gemma-2-9B-It}: Gemma Terms of Use \citep{gemma2report2024}.
\item \textbf{TaskTracker} probe code and dataset: open-sourced by
  \citet{tasktracker2025}; we used their probe protocol on our pair
  set without redistributing their dataset.
\item \textbf{AgentDojo}: Apache-2.0 \citep{agentdojo2024}.
\item \textbf{TransformerLens} \citep{nanda2022transformerlens}: MIT.
\item \textbf{nnsight} \citep{fiotto2024nnsight}: MIT.
\end{itemize}

\section{Statistical Procedures}
\label{app:bootstrap}

For ASR and utility in Tables~\ref{tab:defense},
\ref{tab:agentdojo-expanded-app}, and~\ref{tab:sampling-app}, we draw
$10{,}000$ percentile-bootstrap resamples with seed \texttt{20260427}. Each
resample draws trajectories with replacement and recomputes the rate. Paired
comparisons resample the same trajectory indices across conditions. The 2.5th
and 97.5th percentiles form the reported interval. Patching standard deviations
are computed across $100$ matched pairs; random-direction summaries use the
mean and standard deviation across $30$ directions. The targeted position comparisons use a two-sided exact McNemar test.
If $b$ baseline successes become failures and $c$ failures become successes,
the test conditions on $b+c$ discordant pairs. For the reported position
tests, $c=0$, giving $p=\min(1,2^{1-b})$.

We calculate the position-test CIs directly from the empirical
paired-bootstrap distribution. With $n=96$ and $c=0$, a resample contains
$K\sim\mathrm{Binomial}(96,b/96)$ improved outcomes and has ASR difference
$-K/96$. The 2.5th and 97.5th percentiles give the intervals in
Tables~\ref{tab:qwen-positions-app} and~\ref{tab:llama-positions-app}.
This exact evaluation of the bootstrap distribution avoids Monte Carlo
variation. The larger AgentDojo evaluation retains its recorded
paired-bootstrap tests. Utility intervals resample benign tasks.

\section{AgentDojo Direction Construction}
\label{app:agentdojo-direction}

We estimate behavior-associated directions $\mathbf{d}_{\text{slack}}$ and
$\mathbf{d}_{\text{workspace}}$ from their respective AgentDojo trajectories.
A trajectory is
\emph{hijacked} if the agent executes the adversarial instruction and
\emph{resisted} if it ignores the instruction and completes the benign task.
At L20, we extract the residual at the last token of the complete serialized
input (the pre-action token) and define the unit direction as
\[
\mathbf{d}_{\text{suite}} =
\frac{\mu_{\text{hijacked}}-\mu_{\text{resisted}}}
{\|\mu_{\text{hijacked}}-\mu_{\text{resisted}}\|_2}.
\]
This contrast is indexed by behavior, not by source role; it is distinct from
$\mathbf{d}_{\mathrm{role}}$ estimated from counterfactual role pairs.
The sample-efficiency analysis uses a $240$-trajectory training pool,
separate from the $240$-trajectory evaluation set in each suite. It draws
$n_{\text{train}}$ trajectories from the training pool and compares the
resulting direction with the full-pool estimate. Layer sweeps and the
synthetic, random, and out-of-band controls test suite specificity in both
channels.

\section{Direction Sample Efficiency}
\label{app:sample-efficiency}

Table~\ref{tab:sample-efficiency} measures how many Slack trajectories are
needed to estimate a useful direction. For each $n_{\text{train}}$, we draw
five subsets without replacement from the $240$ training trajectories. We estimate
$\mathbf{d}_{\text{slack}}$ on each subset, then measure its cosine with the
full-data direction and its held-out ASR reduction. With $32$ trajectories,
the cosine reaches $0.852 \pm 0.041$ and ASR falls by $9.8 \pm 1.5$ points,
or $73\%$ of the full-data effect.

\begin{table}[!t]
\caption{Sample-efficiency sweep for $\mathbf{d}_{\text{slack}}$
  (Qwen-2.5-7B, L20, $\lambda=32$). Mean and standard deviation
  across five random subset seeds for $n_{\text{train}}<240$.
The full-pool endpoint uses one fixed direction, so no subset SD is reported.}
\label{tab:sample-efficiency}
\centering\small
\begin{tabular}{lccc}
\toprule
$n_{\text{train}}$ & Cosine to full $\mathbf{d}$ & ASR change (pp) & Utility drop (pp) \\
\midrule
$8$   & $0.452 \pm 0.185$ & $-2.1 \pm 1.8$ & $+5.2 \pm 3.4$ \\
$16$  & $0.684 \pm 0.092$ & $-5.4 \pm 2.2$ & $+2.8 \pm 1.5$ \\
$32$  & $0.852 \pm 0.041$ & $-9.8 \pm 1.5$ & $+1.8 \pm 0.8$ \\
$64$  & $0.925 \pm 0.022$ & $-11.5 \pm 0.8$ & $+1.5 \pm 0.4$ \\
$128$ & $0.968 \pm 0.012$ & $-12.8 \pm 0.5$ & $+1.6 \pm 0.3$ \\
$240$ & $1.000$ & $-13.4$ & $+1.6$ \\
\bottomrule
\end{tabular}
\end{table}

\section{Bidirectional Patching on AgentDojo Trajectories}
\label{app:ipi-patching}

We form matched pairs from AgentDojo-Slack. Each pair shares the benign task
and tool-output frame, but one trajectory is hijacked and the other resisted.
Following \S\ref{sec:patching}, we patch in both directions from L4 to L24. At
the pre-action token (the last token of the complete serialized input),
the source residual replaces the target residual
at the same position. For resisted recipients at L20, ASR remains low
under a random direction ($5.6\%$), a same-layer random residual ($6.1\%$),
or a non-final-position patch ($7.5\%$).

\begin{figure}[!t]
\centering
\includegraphics[width=0.85\textwidth]{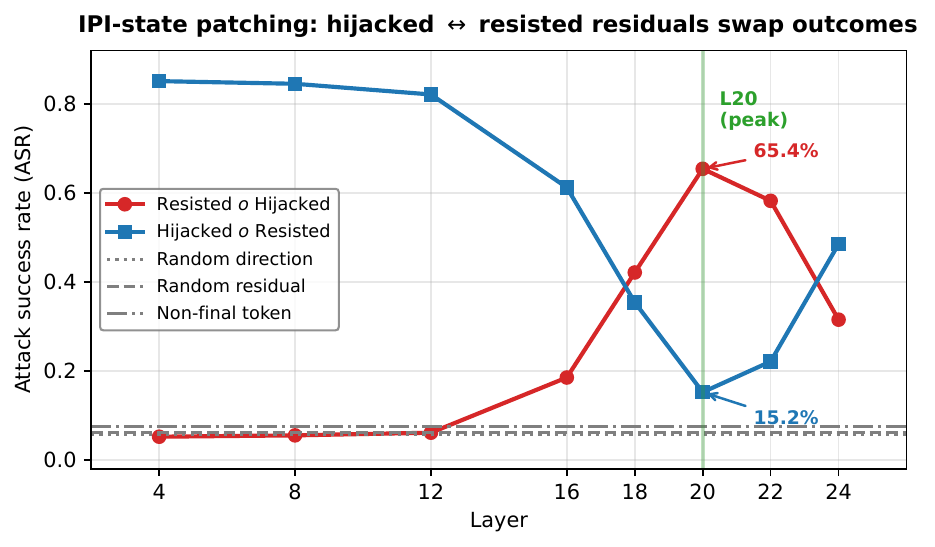}
\caption{Bidirectional state patching on the matched Qwen-2.5-7B
AgentDojo-Slack trajectories. Pre-action swaps have their largest effect
at L18--L22; ASR moves back toward the unpatched outcome at L24.}
\label{fig:ipi-patching}
\end{figure}

\section{Tool-Argument Injection}
\label{app:tool-argument}

The tool-argument evaluation uses function-calling JSON for six tools: weather,
calculator, email, calendar, search, and file access. Each prompt pairs a
benign request with a tool output that contains an adversarial argument, such
as ``send email to attacker@example.com.'' An attack succeeds when the model
places that argument in its JSON call. On clean tool outputs, benign accuracy
checks whether the model supplies the correct argument. The L24 directional
intervention with $\lambda=32$ lowers malicious-argument use from
$88.2\%$ to $12.4\%$.
Benign accuracy changes from $92.5\%$ to $91.2\%$, and JSON validity from
$98.1\%$ to $97.6\%$.

\clearpage
\section{Train--Test Separation for Direction Estimation}
\label{app:split-audit}

Table~\ref{tab:split-audit} summarizes train--test separation. The synthetic
tool-output split holds out content and trajectories, but only some attack
families; it supports localization and operating-point selection. AgentDojo
and memory directions use disjoint trajectories, and their leave-family-out
splits measure transfer to unseen attacks.

\begin{table}[!t]
\caption{Train--test separation for direction estimation. The last three
columns indicate whether the corresponding unit is held out.}
\label{tab:split-audit}
\centering\small
\begin{tabular}{lccccc}
\toprule
Direction & Training source & Evaluation source & Content & Family & Trajectory \\
\midrule
$\mathbf{d}_{\text{tool}}$ & Synthetic role pairs & Synthetic attacks & Yes & Partial & Yes \\
$\mathbf{d}_{\text{slack}}$ & Slack train (LAF-out) & Slack held-out & Yes & Yes & Yes \\
$\mathbf{d}_{\text{workspace}}$ & Workspace train (LAF-out) & Workspace eval & Yes & Yes & Yes \\
$\mathbf{d}_{\text{memory}}$ & Memory role pairs & Memory poisoning & Yes & Yes & Yes \\
\bottomrule
\end{tabular}
\end{table}

\begin{table}[!t]
\caption{Rank-$k$ SVD intervention on AgentDojo (Qwen-2.5-7B, L20,
  $\lambda=32$).}
\label{tab:rank-k}
\centering\small
\begin{tabular}{lcc}
\toprule
Intervention & Slack ASR & Workspace ASR \\
\midrule
No intervention & 34.6\% & 31.2\% \\
Rank-1 (in-channel) & 21.8\% & 20.2\% \\
Rank-2 (in-channel) & 20.4\% & 19.1\% \\
Rank-4 (in-channel) & 19.8\% & 18.9\% \\
Rank-8 (in-channel) & 19.5\% & 18.8\% \\
\midrule
Rank-1 Slack $\to$ Workspace & --- & 30.5\% \\
Rank-1 Workspace $\to$ Slack & 32.2\% & --- \\
\bottomrule
\end{tabular}
\end{table}

\section{Low-Rank Intervention Analysis}
\label{app:rank-k}

We apply SVD to matched hijacked-minus-resisted residual differences at Slack
L20. Rank-$k$ interventions show diminishing returns
(Table~\ref{tab:rank-k}). Rank 1 lowers ASR from $34.6\%$ to $21.8\%$; ranks
2, 4, and 8 reach $20.4\%$, $19.8\%$, and $19.5\%$. Cross-channel rank-1
directions transfer poorly: Slack$\to$Workspace yields $30.5\%$ ASR, and
Workspace$\to$Slack yields $32.2\%$, both near baseline. Most of the measured within-suite ASR reduction is obtained at low rank,
but the learned rank-1 direction does not transfer effectively to the other
suite. Table~\ref{tab:rank-k} uses the SVD rank-1 vector;
Figure~\ref{fig:expanded-agentdojo} uses a mean-difference direction. The two
rank-1 results therefore use different estimators.

\end{document}